\pdfoutput=1
\documentclass[11pt]{article}

\usepackage[final]{acl}

\usepackage{times}
\usepackage{latexsym}

\usepackage[T1]{fontenc}
\usepackage[utf8]{inputenc}

\usepackage{microtype}

\usepackage{inconsolata}

\usepackage{graphicx}

\usepackage{threeparttable}
\usepackage{booktabs}
\usepackage{multirow}
\usepackage{multicol}
\usepackage{amsmath}
\usepackage{amssymb} 
\usepackage[ruled,vlined]{algorithm2e}
\usepackage{subcaption}
\title{EXCEEDS: Extracting Complex Events via Nugget-based Grid Modeling in Scientific Domain}

\author{
 \textbf{Yi-Fan Lu}$^1$,
 \textbf{Xian-Ling Mao$^1$\thanks{Corresponding Author.}},
 \textbf{Bo Wang}$^1$,
 \textbf{Xiao Liu}$^2$,
 \textbf{Heyan Huang}$^1$
\\
 $^1$Beijing Institute of Technology, $^2$Microsoft Research Asia
\\
\texttt{\href{mailto:yifanlu@bit.edu.cn}{yifanlu@bit.edu.cn}},
\texttt{\href{mailto:maoxl@bit.edu.cn}{maoxl@bit.edu.cn}}, 
\texttt{\href{mailto:bwang@bit.edu.cn}{bwang@bit.edu.cn}}, \\
\texttt{\href{mailto:xiaoliu@bit.edu.cn}{xiaoliu2@microsoft.com}}, 
\texttt{\href{mailto:hhy63@bit.edu.cn}{hhy63@bit.edu.cn}}
}

\begin{document}
\maketitle
\begin{abstract}

It is crucial to understand a specific domain by events. 
Extensive event extraction research has been conducted in many domains such as news, finance, and biology.
However, event extraction in scientific domain is still insufficiently supported by comprehensive datasets and tailored methods.
Compared with other domains, scientific domain has two characteristics: (1) denser nuggets and events, and (2) more complex information forms.
To solve the above problem, considering these two characteristics, we first construct SciEvents, a large-scale multi-event document-level dataset with a schema tailored for scientific domain. 
It consists of 2,508 documents and 24,381 events under multi-stage manual annotation and quality control. 
Then, we propose EXCEEDS, an end-to-end scientific event extraction framework by encoding dense nuggets into a grid matrix and simplifying complex event extraction as a nugget-based grid modeling task.
Experiments on SciEvents demonstrate state-of-the-art performances of EXCEEDS. 
Both the SciEvents dataset and the EXCEEDS framework are released publicly to facilitate future research.\footnote{https://github.com/HammerScholar/EXCEEDS}

\end{abstract}

\section{Introduction}
% 简单介绍EE这个任务
Event extraction (EE) is a fundamental information extraction task aiming to extract structural
event knowledge from plain texts \cite{peng-etal-2023-devil}. 
% It aims to detect event instances as well as all event-related participants and attributes in texts \cite{8918013, 7243219}. 
It is typically decomposed into two pipeline subtasks: event detection (ED) and event argument extraction (EAE). 
Specifically, ED identifies a word span (hereafter referred to as a \textbf{nugget}) that most clearly refers to the occurrence of an event, \textit{i.e.}, event trigger, and also detects the event type evoked by the event trigger \cite{pouran-ben-veyseh-etal-2022-minion}. 
Given an event trigger and its event type, EAE further identifies nuggets as event arguments and classifies their roles in the event. 

\begin{figure}[ht]
  \centering
  \includegraphics[width=\linewidth]{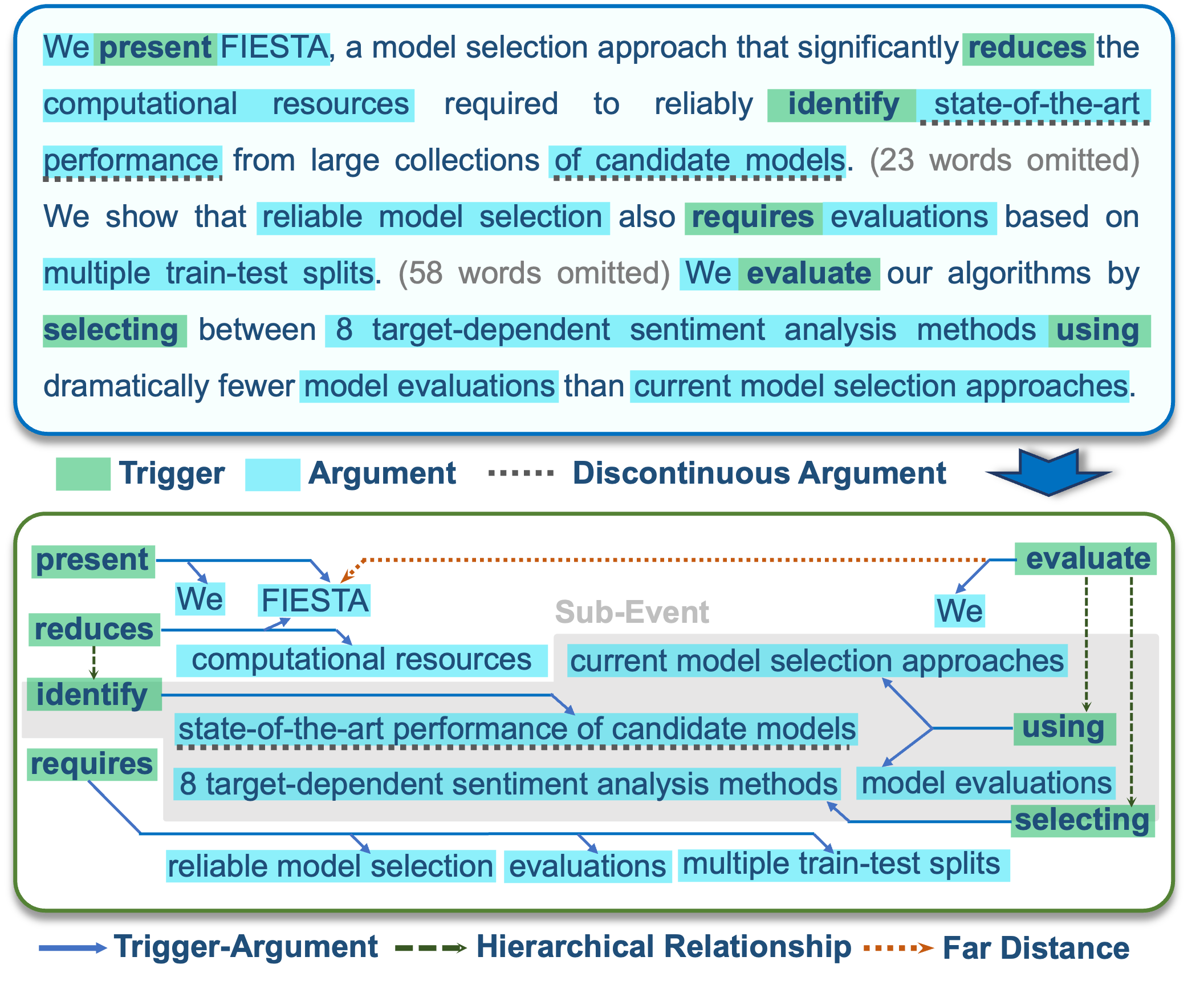}
  \caption{A real example from SciEvents. The upper panel displays a scientific paper abstract, and the lower panel shows the extracted events, highlighting the dense information and complex event structures characteristic of scientific texts.}
  \label{fig:example}
\end{figure}

%%% It is crucial to utilize events to understand a specific domain. Extensive event extraction research has been conducted in many domains such as news, finance, and biology.
EE provides an effective abstraction for representing domain knowledge and supports downstream tasks such as reasoning, summarization, and knowledge discovery.
Consequently, extensive EE research has been conducted in various scenarios and domains such as
internet news, radio conversations, internet blogs \cite{sundheim-1992-overview, aguilar-etal-2014-comparison, ebner-etal-2020-multi},   % MUC-4, ACE2005, RichERE, TAC-KBP, RAMS, M2E2
business \cite{yang-etal-2018-dcfee, ijcai2020p0619},  % DCFEE, GNBusniess, ChFinAnn, FMC
biology \cite{kim-etal-2011-overview-genia, pyysalo2012event},  % genia2011, genia2013, mlee
legislation \cite{shen-etal-2020-hierarchical, yao-etal-2022-leven},  % unknown name, unknown name, LEVEN
cybersecurity \cite{man-duc-trong-etal-2020-introducing, Satyapanich_Ferraro_Finin_2020}  % CySecED, CASIE
and so on.
Despite this progress, scientific literature has been growing rapidly in recent decades, with millions of new publications released every year.
Such growth poses an urgent challenge for managing scientific domain knowledge, calling for effective EE solutions.

%%% However, scientific domain still lacks event extraction research, including comprehensive datasets and corresponding methods. Compared with other domains, scientific domain presents two characteristics: denser nuggets and more complex events.
However, EE in scientific domain remains insufficiently characterized by existing datasets and methods.
In particular, current resources and formulations struggle to capture two salient characteristics of scientific texts.
\textbf{First}, compared with other domains, scientific domain tends to contain \textbf{more complex information forms}. 
Although many EE methods are task-specialized and rely on domain-specific ontologies \cite{lu-etal-2022-unified}, these ontologies typically adopt flat tabular schema, which (1) neglects the hierarchical structure of events, (2) restricts the continuity of arguments, and (3) complicates the coreference problem, while these complex information forms are common in scientific literature. 
For example in Figure~\ref{fig:example}, (1) the trigger \textit{evaluate} and the trigger \textit{using} form a hierarchical relationship; (2) the trigger \textit{identity} connects a discontinuous argument \textit{state-of-the-art performance of candidate models}; (3) the trigger \textit{evaluate} connects an argument \textit{FIESTA} with a far distance from itself. 
These three examples demonstrate complex nuggets and events in scientific domain.
\textbf{Second}, scientific texts, especially literature abstracts, tend to contain \textbf{denser nuggets and events} (with statistical evidence presented in Section~\ref{sec:statistics_analysis}). 
Unlike many existing datasets that focus on sentence-level extraction or single-event documents, scientific domain EE requires modeling dense, document-level multi-event interactions (see Figure~\ref{fig:example} and Section~\ref{sec:statistics_analysis}).
Together, these two characteristics motivate the need for dedicated EE resources and methods for scientific domain.

%%% To solve the above problem, considering these two characteristics, we first construct SciEvents, a large-scale multi-event document-level dataset with a schema tailored for scientific domain. 
To explore the unique characteristics of scientific domain EE, we first introduce SciEvents, a large-scale document-level multi-event dataset tailored for scientific literature.
SciEvents contains 2,508 manually annotated abstracts with 24,381 events, and is constructed with a refined schema designed to capture dense and structurally complex event patterns.
Dataset statistics show that SciEvents exhibits both \textbf{denser nugget distributions} and \textbf{more complex event structures} than existing domain-specific EE datasets, reflecting the information-intensive nature of scientific texts.
% \textbf{Then}, we further propose EXCEEDS, a framework to \textbf{ex}tract \textbf{c}omplex \textbf{e}vents via nugg\textbf{e}t-based gri\textbf{d} modeling in \textbf{s}cientific domain. 
% EXCEEDS represents pairwise token relations across the entire document in a word-word event grid, enabling unified modeling of dense multi-event contexts as well as complex nugget and event structures, including hierarchical relations, discontinuous arguments, and long-distance dependencies. This formulation allows EXCEEDS to effectively address the challenges posed by scientific texts under a single end-to-end framework.

Denser nuggets and more complex events pose two fundamental challenges to existing EE methods. 
On the one hand, the high density requires models to capture global information, rather than extracting events only at the sentence level or as isolated instances. 
On the other hand, the complexity of scientific events calls for models that can represent hierarchical relationships, handle discontinuous nuggets, and associate triggers with arguments across long textual distances.
However, most existing EE approaches are developed under assumptions of non-hierarchical structures and locally bounded contexts, which limit their effectiveness in modeling the complex event patterns commonly observed in scientific texts.
% This motivates the need for a more unified formulation that can capture fine-grained token dependencies at the document level.  % 新加的一句

To address these challenges, we further propose \textbf{EXCEEDS}, an end-to-end framework to \textbf{EX}tract \textbf{C}omplex \textbf{E}vents via nugg\textbf{E}t-based gri\textbf{D} modeling in \textbf{S}cientific domain. 
EXCEEDS represents pairwise token relations across the entire document in a word-word event grid, enabling unified modeling of dense multi-event contexts as well as complex nugget and event structures, including hierarchical relations, discontinuous arguments, and long-distance dependencies. This formulation allows EXCEEDS to effectively address the challenges posed by scientific texts under a single end-to-end framework.

We evaluate state-of-the-art and recent EE methods on SciEvents under an extended evaluation protocol that incorporates an event correlation metric for hierarchical EE. Experimental results show that EXCEEDS achieves consistently strong performance across tasks, while further analysis reveals that complex nugget structures, especially under dense scientific contexts, remain challenging for existing models.

%%% Experiments on SciEvents demonstrate state-of-the-art performances of EXCEEDS. 
% TODO: 这里最后一点补一下实验发现
% our contributions are three-fold：（1）提出了一个method，能够有效解决复杂EE（2）提出了一个science domain的数据集，该数据集很好地体现了science domain的特点，并有独到之处，能够推动science information extraction的发展（3）method在新提出的数据集上有很好的表现
In summary, our contributions are two-fold: 
(1) We introduce SciEvents, a large-scale document-level EE dataset for the scientific domain with a refined schema, providing a comprehensive benchmark for studying dense and complex scientific events.
% (1) To address the lack of event extraction research in scientific domain, we construct SciEvents, a large-scale event extraction dataset in scientific domain with a refined schema. The proposed dataset is the first scientific event extraction dataset.
(2) We propose EXCEEDS, an end-to-end EE framework tailored to the challenges of density and structural complexity in scientific texts, which achieves state-of-the-art performance on SciEvents.
% by modeling pairwise token relations in a grid structure and simplifying complex event extraction into a dot construction and connection task.
% (2) To solve problems of dense and complex nuggets and events in scientific domain, we propose EXCEEDS, a novel end-to-end extractive method to model relationships of all token pairs at once and simplify complex event extraction into a dot construction and connection task.
% (3) We extend the event extraction task on SciEvents with a complementary event correlation metric for hierarchical events. Extensive experiments demonstrate state-of-the-art performance of the proposed method EXCEEDS on SciEvents.
% show that EXCEEDS achieves state-of-the-art performance across all tasks, with notable improvements in hierarchical event extraction.
% (3) We define the event extraction task on SciEvents with additional complementary metrics. Experiments demonstrate state-of-the-art performance of the proposed method EXCEEDS on SciEvents.

\begin{table*}
    \small
    \centering
    \begin{tabular}{l|lrrrrrr}
    \toprule
        \bf Dataset & Domain & \#Docs & \#Tokens & \#ETs & \#Events & \#ATs & \#Arguments \\
    \midrule
        ACE2005 \cite{doddington-etal-2004-automatic} & News & 599 & 297,842 & 33 & 5,348 & 22 & 8,097 \\
        Genia2011 \cite{kim-etal-2011-overview-genia} & Biomedical & 1,375 & 345,371 & 9 & 13,537 & 10 & 11,865 \\
        Genia2013 \cite{kim-etal-2013-genia} & Biomedical & 20 & 149,856 & 13 & 6,001 & 7 & 5,660 \\
        RAMS \cite{ebner-etal-2020-multi} & News & 9,124 & 1,218,622 & 139 & 9,124 & 65 & 21,237 \\
        CASIE \cite{Satyapanich_Ferraro_Finin_2020} & Cybersecurity & 999 & 387,275 & 5 & 8,479 & 26 & 22,679 \\
        M2E2 \cite{li-etal-2020-cross} & Multimedia & 6,013 & 169,990 & 8 & 1,105 & 15 & 1,659 \\
        WikiEvents \cite{li-etal-2021-document} & News & 246 & 189,718 & 50 & 3,951 & 59 & 5,536 \\
        PHEE \cite{sun-etal-2022-phee} & Drug Safety & 4,827 & 106,447 & 2 & 5,019 & 16 & 25,760 \\
        % DocEE \cite{tong-etal-2022-docee} & Wikipedia & 27,485 & 16,117,229 & 59 & 27,485 & 356 & 405,584 \\
        % GENEVA \cite{parekh-etal-2023-geneva} & General & 97 & 98,480 & 115 & 7,505 & 220 & 12,269 \\
        Maccrobat-EE \cite{ma-etal-2023-dice} & Clinical & 200 & 107,130 & 14 & 13,128 & 22 & 8,599 \\
        % MAVEN-Arg \cite{wang-etal-2024-maven} & General & 4,480 & 1,149,539 & 162 & 80,479 & 143 & 236,937 \\
    \midrule 
        SciEvents (Ours) & Science & 2,508 & 439,890 & 10 & 24,381 & 20 & 56,411 \\
    \bottomrule
    \end{tabular}
    \caption{Basic statistics of widely-used domain-specific event datasets. This table only presents publicly available event datasets that include argument annotations. \#ETs: number of event types. \#ATs: number of argument types.}
    \label{tab:basic_statistics}
\end{table*}

\section{Related Works}

\paragraph{Event Extraction Datasets}
EE datasets have been constructed across a wide range of domains.
Early efforts mainly focus on the news domain, describing events in realistic scenarios \cite{ace2005, aguilar-etal-2014-comparison, song-etal-2015-light}. 
General domain datasets are further built from diverse sources like Wikipedia \cite{wang-etal-2020-maven, li-etal-2021-document, tong-etal-2022-docee}, Reddit \cite{ebner-etal-2020-multi}, Baidu news \cite{li2020duee}, FrameNet \cite{parekh-etal-2023-geneva} and multi-lingual candidate data \cite{pouran-ben-veyseh-etal-2022-minion}. 
In addition, domain-specific datasets have been developed for finance \cite{yang-etal-2018-dcfee, liu-etal-2019-open}, biomedicine and related fields \cite{kim-etal-2011-overview-genia, sun-etal-2022-phee, ma-etal-2023-dice}, as well as other domains such as cybersecurity, law, and literature \cite{man-duc-trong-etal-2020-introducing, shen-etal-2020-hierarchical, sims-etal-2019-literary}.
Despite these advances, event datasets for the scientific domain remain limited, and existing resources rarely analyze the characteristics of scientific texts. 
In this work, we systematically examine the characteristics of scientific abstracts and construct SciEvents, a document-level EE dataset tailored to the scientific domain.

\paragraph{Event Extraction Approaches}
EE has evolved from early sequence labeling methods to more advanced neural architectures.
To jointly model heterogeneous elements in EE datasets \cite{peng-etal-2023-devil}, early work focuses on joint extraction frameworks that capture dependencies within and across events \cite{liu-etal-2018-jointly, yang2019exploring, nguyen-etal-2021-cross, lin-etal-2020-joint}.
Subsequent studies reformulate EE as machine reading comprehension, enabling more flexible trigger and argument extraction via question answering \citep{chen-etal-2020-reading, li2020event, zhou2021role, wei2021trigger}. 
More recent approaches adopt sequence-to-structure generation with Transformer-based models, unifying ED and EAE within a single framework \citep{lu-etal-2021-text2event, Lou_Lu_Dai_Jia_Lin_Han_Sun_Wu_2023,wang-etal-2023-boosting, liu-etal-2022-dynamic, yang-etal-2024-scented}. 
With the emergence of large language models (LLMs), EE has further benefited from strong generalization and zero-shot capabilities \citep{wei2023zero, gao2023exploring, wang2023instructuie, sainz2024gollie, gao2023benchmarking, li-etal-2024-knowcoder}. 
Despite these advances, many existing methods struggle to model structurally complex event mentions.
Some work partially mitigates this problem by modeling token-level relations \cite{Lou_Lu_Dai_Jia_Lin_Han_Sun_Wu_2023, liu-etal-2023-rexuie, zhu-etal-2023-mirror}, or by adopting a more universal information extraction paradigm \cite{lu-etal-2022-unified, li-etal-2024-knowcoder}. 
However, these approaches either rely on span-boundary representations \cite{Lou_Lu_Dai_Jia_Lin_Han_Sun_Wu_2023, liu-etal-2023-rexuie}, require instruction-style inputs with schema conditioning \cite{zhu-etal-2023-mirror}, or rely on multi-task and multi-dataset training \cite{lu-etal-2022-unified, li-etal-2024-knowcoder}. 
In this paper, we propose EXCEEDS, an end-to-end pairwise token relation modeling framework over the entire document, using only raw text as input and targeting the EE-only, single-dataset setting.

% Another line of work formulates EE within a universal information extraction framework \cite{lu-etal-2022-unified, Lou_Lu_Dai_Jia_Lin_Han_Sun_Wu_2023, liu-etal-2023-rexuie, zhu-etal-2023-mirror, wang2023instructuie, li-etal-2024-knowcoder}. However, they rely on multi-task and multi-dataset training, which are not directly aligned with the EE-only, single-dataset setting in the work.

\section{The SciEvents Dataset}
To support systematic research on scientific EE, we construct SciEvents, a large-scale document-level EE dataset tailored for scientific literature.
In this section, we will introduce the dataset construction process in Section~\ref{sec:construction_process}, and present a comprehensive statistical analysis in Section~\ref{sec:statistics_analysis}.

\begin{table*}
    \small
    \centering
    \begin{tabular}{ll|rrrrrrr}
    \toprule
        \multirow{2}{*}{\bf Dataset} & \multirow{2}{*}{Domain} & \multicolumn{3}{c}{Density: Every 100 Tokens Contains} & \multicolumn{4}{c}{Complexity: Complex Forms} \\
    \cmidrule(lr){3-5}
    \cmidrule(lr){6-9}
        ~ & ~ & \#Events & \#Args & \#Nugget Tokens & \#D(\%) & \#O(\%) & \#R(\%) & \#S(\%) \\
    \midrule
        ACE2005 & News & 1.80 & 2.72 & 4.62 & -- & 13.88 & -- & -- \\
        Genia2011 & Biomedical & 3.92  & 3.44 & 9.31 & -- & 36.62 & -- & -- \\
        Genia2013 & Biomedical & 4.00  & 3.78 & 8.58 & -- & 34.00 & -- & -- \\
        RAMS & News & 0.75  & 1.74 & 4.06 & -- & 10.51 & -- & -- \\
        CASIE & Cybersecurity & 2.19  & 5.86 & 17.80 & -- & 1.40 & -- & -- \\
        M2E2 & Multimedia & 0.65  & 0.98 & 1.89 & -- & 6.08 & -- & -- \\
        WikiEvents & News & 2.08  & 2.92 & 5.69 & -- & 8.19 & -- & -- \\
        PHEE & Drug Safety & 4.72 & 24.20 & 53.43 & -- & 64.03 & -- & -- \\
        % DocEE & Wikipedia & 0.17  & 2.52 & 7.40 & -- & 11.16 & -- & -- \\
        % GENEVA & General & 7.62  & 12.46 & 52.15 & -- & 46.43 & -- & 1.29 \\
        Maccrobat-EE & Clinical & 12.25  & 8.03 & 38.65 & -- & 3.71 & -- & -- \\
        % MAVEN-Arg & General & 7.00  & 20.61 & 33.09 & -- & 68.42 & -- & 2.29 \\
    \midrule
        SciEvents (Ours) & Science & 5.54 & 12.82 & 39.49 & 3.08 & 33.70 & 1.01 & 25.63 \\
    \bottomrule
    \end{tabular}
    \caption{Statistics of density and complexity of widely-used domain-specific event datasets. This table only presents publicly available event datasets that include argument annotations. \#D: Discontinuous nugget. \#O: Overlapping nugget. \#R: Reverse-order nugget. \#S: Sub-event.}
    \label{tab:dense_complex_statistics}
\end{table*}

\subsection{Dataset Construction Process}
\label{sec:construction_process}
\paragraph{Schema Design}
Scientific abstracts are typically organized around four rhetorical components: \emph{background}, \emph{related work}, \emph{methodology}, and \emph{results}.
Motivated by this regularity, we design an event schema comprising 10 event types that cover these components.
For instance, abstracts often summarize prior approaches and highlight their limitations; we capture such information using the \textit{RelatedWorkStep} and \textit{RelatedWorkFault} event types, respectively.

For each event type, we define a set of argument types to encode the information that readers typically seek in scientific abstracts.
To ensure both coverage and annotation feasibility, we develop the schema iteratively with domain experts. Specifically, two professors and three senior Ph.D. students in computer science annotate a set of seed documents and revise the schema over four rounds.
The final schema and detailed definitions of all event types and argument types are provided in Appendix~\ref{sec:schema}.
We collect papers from the recent 4 years (2019-2022) ACL main conference paper abstracts as candidate data.

\paragraph{Annotation and Quality Control}
During the pre-annotation stage, we train three supervisors and some candidates, resulting in seven qualified annotators for the official annotation.
For reproducibility, detailed descriptions of the annotation protocol are provided in Appendix~\ref{sec:construction_details}.

Quality inspection is conducted by three supervisors and two well-performing annotators.
The annotator and the inspector of a document are strictly separated.
If a document contains more than two annotation conflicts (including missing annotations), it is returned to the original annotator together with detailed revision comments provided by a quality inspector.
Otherwise, minor conflicts are corrected by the inspection team, and corresponding feedback is still provided to annotators to facilitate continuous improvement.
The first-pass inspection acceptance rate is 73.05\%.
A fully annotated example document can be found in Appendix~\ref{sec:annotaion_example}.

\subsection{Dataset Statistics Analysis}
\label{sec:statistics_analysis}
This section will provide a comprehensive statistical analysis of SciEvents, with particular emphasis on information density and complexity.

\paragraph{Basic Statistics}
Table~\ref{tab:basic_statistics} presents basic statistics of SciEvents and other widely-used domain-specific EE datasets covering diverse domains. 
Among these datasets, SciEvents is distinguished as a large-scale dataset specifically constructed for the scientific domain.
In terms of annotation scale, SciEvents 24,381 event instances and 56,411 arguments, substantially exceeding most existing domain-specific event datasets.
Notably, SciEvents achieves this scale with only 10 event types. This suggests that the large number of event instances in SciEvents primarily arises from frequent event occurrences within scientific documents, rather than an expanded or fine-grained schema, reflecting the information-intensive nature of scientific texts.
Statistics can be found in Appendix~\ref{sec:scievents_distribution}.
%%% 说不定后面rebuttal能用上
% Notably, SciEvents achieves this scale with a substantially more compact event schema. While DocEE and MAVEN-Arg define 59 and 162 event types respectively, SciEvents contains only 10 event types yet includes 24,381 event instances.

\paragraph{Information Density Statistics}
As shown in Table~\ref{tab:dense_complex_statistics}, SciEvents exhibits high information density under all token-normalized metrics, with 5.54 events, 12.82 arguments, and 39.49 nugget tokens per 100 tokens, indicating that a large proportion of tokens in scientific documents directly participate in event expressions.

Among other domain-specific datasets, similarly high density values are mainly observed in medical datasets such as PHEE and Maccrobat-EE, whose documents describe inherently information-intensive content (\textit{e.g.}, drug safety reports and clinical records). By contrast, remaining datasets generally exhibit substantially lower densities. Overall, these results suggest that the elevated density of SciEvents reflects intrinsic properties of scientific texts under domain-specific settings.

%%% 包含GENEVA和MAVEN-Arg的版本，说不定后面rebuttal能用上
% Although general-domain datasets such as GENEVA and MAVEN-Arg report even higher event and argument densities, these values are partly attributable to their substantially larger and more fine-grained event and argument schemas, which naturally increase instance frequency per unit text.

% Among domain-specific datasets, SciEvents remains one of the densest resources, comparable to high-density biomedical and clinical datasets such as PHEE and Maccrobat-EE, whose documents similarly describe information-intensive phenomena (\textit{e.g.}, drug adverse events and clinical diagnoses). Moreover, SciEvents achieves a notably high nugget token density, suggesting that a large proportion of tokens in scientific texts directly contribute to event expressions. Together, these results confirm that the elevated density of SciEvents reflects intrinsic properties of scientific writing rather than schema-induced artifacts.

\paragraph{Information Complexity Statistics}
The right part of Table~\ref{tab:dense_complex_statistics} reports the proportions of complex event forms. Unlike most existing domain-specific datasets that mainly annotate contiguous nuggets, SciEvents explicitly covers diverse complex structures, including overlapping, discontinuous, reverse-order nuggets, and sub-events.
Specifically, SciEvents contains a substantial proportion of overlapping nuggets (33.70\%) and sub-events (25.63\%), together with non-negligible occurrences of discontinuous (3.08\%) and reverse-order nuggets (1.01\%). By contrast, other datasets only partially capture overlapping structures. These statistics reflect the structural complexity of scientific events and highlight the challenges they pose for EE models.

\section{Event Extraction Problem Formulation}
In domain-specific EE, a predefined schema is given as $ S = \{ T_E, T_A \} $, where $T_E$ denotes an event type set and $T_A$ denotes an argument type set.
Each event type $ t_e \in T_E $ is associated with a specific argument type set $T_A(t_e)$. 
Given a document $D $, EE aims to extract a set of events $E = \{e_1, e_2, \ldots, e_M\} $ in $D$, where each event $e = \{t_e, t, A \}$ consists of an event type $ t_e \in T_E $, a trigger $t$ and a set of arguments $ A = \{ a_{1}, a_{2}, \ldots, a_N \} $. 
Each argument $ a = \{ t_a, m \}$ consists of an argument type $ t_a \in T_A(t_e) $ and a word span $ m $. 
Both triggers and arguments are referred to as nuggets, whose word spans should be combinations of tokens in $ D $. 
In SciEvents, we add an event correlation task to extract hierarchical event structures. Specifically, the trigger $t_s$ of a sub-event $e_s = \{ t_{se}, t_s, A_s \}$ will be regarded as an argument of a main-event $e_m = \{ t_{me}, t_m, A_m \}$ with a certain argument type $t_{sa}$, \textit{i.e.} $\{t_{sa}, t_s\} \in A_m$.

\section{The EXCEEDS Method}
To address the challenges of high density and complexity in scientific EE, we propose EXCEEDS, an end-to-end framework that simplifies EE into a nugget-based grid modeling task.
Section~\ref{sec:grid} introduces the word-word event grid construction,
Section~\ref{sec:framework} presents the overall framework,
and Section~\ref{sec:train_and_inference} describes the training objectives and inference procedure.

\begin{figure}[ht]
  \centering
  \includegraphics[width=\linewidth]{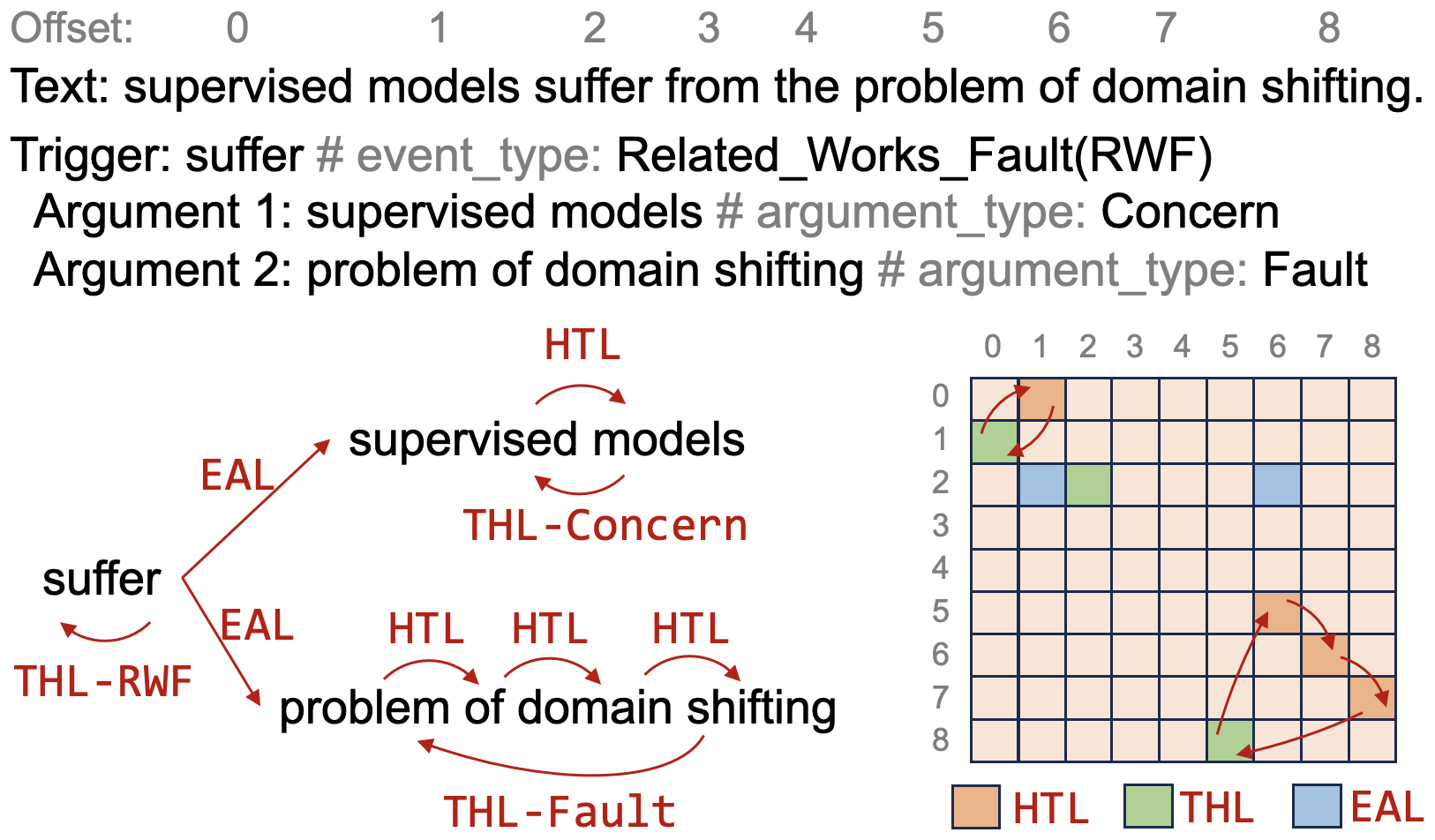}
  \caption{Illustration of the word-word event grid. \textsc{HTL} captures successive token order within a nugget; \textsc{THL} connects the last and first tokens to indicate nugget types; and \textsc{EAL} encodes relations across nuggets. The grid matrix (\textit{right}) presents these relations.}
  \label{fig:grid}
\end{figure}

\subsection{Word-Word Event Grid}
\label{sec:grid}
To effectively capture the complex structures of nuggets and events, we encode relations within a nugget and across different nuggets through a word-word grid.
Formally, given a document $D = \{x_1, x_2, \ldots, x_l\}$, we construct an $l \times l$ grid $G$, where each cell $G[i,j]$ stores the relation type $r \in R$ between token $x_i$ (row) and token $x_j$ (column).
Specifically, within a nugget, we use head-tail-link (\textsc{HTL}) to represent the successive order between adjacent tokens and tail-head-link (\textsc{THL}) to connect the last token back to the first token, which conveys nugget type information.
Across different nuggets, we define event-argument-link (\textsc{EAL}) to represent the relation between a trigger and its argument, or between an event trigger and a sub-event trigger.
For example, Figure~\ref{fig:grid} shows how \textsc{HTL}, \textsc{THL} and \textsc{EAL} are instantiated in SciEvents, and how they are encoded into the corresponding cells of the grid, resulting in a unified representation for nuggets and events.

The benefits of this grid and relation design are threefold. 
First, it enables encoding of complex nugget structures within a document, including overlapping, discontinuous, and reverse-order nuggets. 
Second, it provides a unified formulation of event detection and event argument extraction in an end-to-end manner, allowing the framework to fully leverage contextual information without relying on separate pipeline modules.
Third, it naturally captures hierarchical event relations by encoding relations between trigger pairs.

\begin{figure*}[ht]
  \centering
  \includegraphics[width=\linewidth]{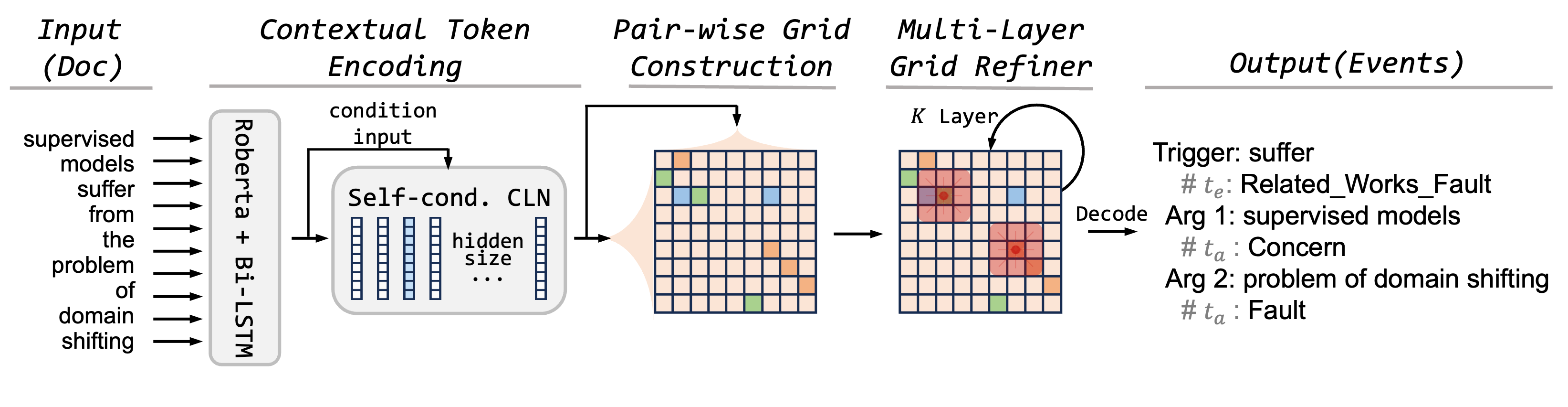}
  \caption{Overall architecture of EXCEEDS. The model encodes contextual token representations, constructs a word-word event grid to model pairwise relations, refines the grid, and decodes events from the refined grid.}
  \label{fig:exceeds}
\end{figure*}

\subsection{The Overall Framework}
\label{sec:framework}

Figure~\ref{fig:exceeds} illustrates the overall architecture of our framework.
Given a document, the model encodes contextual token representations and constructs a word-word grid to jointly model nugget structures and event relations in an end-to-end manner.

\paragraph{Contextual Token Encoding}
We encode the input document using a pretrained language model \cite{liu2019robertarobustlyoptimizedbert} to obtain initial contextual representations, which are further refined by a bidirectional LSTM \cite{huang2015bidirectionallstmcrfmodelssequence} to capture sequential dependencies.
The resulting representations are normalized using Conditional Layer Normalization (CLN) \cite{liu-etal-2021-modulating} to enhance stability and contextual adaptability.
Specifically, given the LSTM outputs $\mathbf{L} \in \mathbb{R}^{l \times d}$, CLN performs normalization with dynamically generated affine parameters conditioned on the contextual representations themselves:
\begin{equation}
\mathbf{H}
=
\text{MLP}_{\gamma}(\mathbf{L})
\odot
\frac{\mathbf{L} - \mu}{\sigma + \epsilon}
+
\text{MLP}_{\beta}(\mathbf{L}),
\end{equation}
where $\mu$ and $\sigma$ denote the mean and standard deviation computed along the feature dimension, $\epsilon$ is a smoothing parameter, and $\odot$ denotes element-wise multiplication.
The output $\mathbf{H} \in \mathbb{R}^{l \times d}$ serves as the contextualized word representations for subsequent pair-wise grid construction.

\paragraph{Pair-wise Grid Construction}
Given the contextualized word representations $\mathbf{H}$, we construct a word-word grid
$\mathbf{G} \in \mathbb{R}^{l \times l \times C_g}$,
where each cell corresponds to relation of a token pair $(x_i, x_j)$.

For each pair, we form a pair-wise representation by concatenating the token representations with a relative distance embedding:
\begin{equation}
\mathbf{z}_{i,j}
=
[\mathbf{h}_i \, ; \, \mathbf{h}_j \, ; \, \mathbf{d}_{i,j}],
\end{equation}
which is projected into the grid feature space via a multilayer perceptron:
\begin{equation}
\mathbf{g}_{i,j}
=
\text{MLP}_{\text{pair}}(\mathbf{z}_{i,j}).
\end{equation}
The resulting $\mathbf{g}_{i,j} \in \mathbb{R}^{C_g}$ constitutes the initial grid representation.

\begin{table*}[ht]
    \small
    \centering
    \begin{threeparttable}
        \begin{tabular}{c|l|ccccc}
            \toprule
             \multicolumn{2}{c}{\bf Model} & TI & TC & AI & AC & EC  \\
             
            \midrule
            \multirow{2}{*}{Global} 
              & OneIE \cite{lin-etal-2020-joint} & \textbf{75.72}$_{\pm0.14}$ & 62.93$_{\pm0.17}$ & 30.30$_{\pm1.85}$ & 28.81$_{\pm1.77}$ & 37.41$_{\pm1.51}$ \\
            ~ & ScentedEAE$^\dagger$ \cite{yang-etal-2024-scented} & 73.27$_{\pm0.52}$ & 63.03$_{\pm0.20}$ & 36.70$_{\pm2.99}$ & 35.74$_{\pm2.41}$ & 37.88$_{\pm2.10}$ \\
            
            \midrule
            % \multirow{4}{*}{Discriminative} 
            ~ & EEQA \cite{du-cardie-2020-event} & 74.85$_{\pm0.78}$ & 62.15$_{\pm0.73}$ & 37.75$_{\pm0.46}$ & 35.64$_{\pm0.59}$ & 44.81$_{\pm1.35}$ \\ % valid
            Discrimi- & PAIE$^\dagger$ \cite{ma-etal-2022-prompt} & 73.27$_{\pm0.52}$ & 63.03$_{\pm0.20}$ & 43.92$_{\pm0.22}$ & 42.06$_{\pm0.34}$ & 47.17$_{\pm0.92}$ \\
            native & Tagprime \cite{hsu-etal-2023-tagprime} & 73.27$_{\pm0.52}$ & 63.03$_{\pm0.20}$ & \underline{44.67}$_{\pm0.13}$ & \underline{42.69}$_{\pm0.32}$ & 47.72$_{\pm0.32}$ \\
            ~ & DEEIA$^\dagger$ \cite{liu-etal-2024-beyond-single} & 73.27$_{\pm0.52}$ & 63.03$_{\pm0.20}$ & 34.86$_{\pm0.81}$ & 33.30$_{\pm0.75}$ & 32.80$_{\pm1.67}$ \\
            
            \midrule
            \multirow{3}{*}{Generative} 
              & BartGen$^\dagger$ \cite{li-etal-2021-document} & 73.27$_{\pm0.52}$ & 63.03$_{\pm0.20}$ & 39.85$_{\pm0.52}$ & 37.81$_{\pm0.39}$ & 42.75$_{\pm0.11}$ \\
            ~ & DEGREE \cite{hsu-etal-2022-degree} & 65.72$_{\pm0.70}$ & 53.52$_{\pm0.75}$ & 28.40$_{\pm0.88}$ & 26.32$_{\pm0.79}$ & 29.53$_{\pm0.96}$ \\
              & KnowCoder \cite{li-etal-2024-knowcoder} & 69.88$_{\pm0.61}$ & 52.02$_{\pm0.34}$ & 35.24$_{\pm0.11}$ & 33.43$_{\pm0.27}$ & 34.54$_{\pm0.82}$ \\
            \midrule
            \multicolumn{2}{l|}{EXCEEDS (Ours)}  & 75.29$_{\pm0.32}$ & \textbf{63.74}$_{\pm0.14}$ & \textbf{44.97}$_{\pm0.28}$ & \textbf{43.20}$_{\pm0.29}$ & \textbf{48.25}$_{\pm0.10}$ \\
            \multicolumn{2}{l|}{~~-- Contextual} & 75.29$_{\pm0.21}$ & \underline{63.44}$_{\pm0.67}$ & 44.07$_{\pm0.51}$ & 42.14$_{\pm0.39}$ & 47.64$_{\pm0.85}$ \\
            \multicolumn{2}{l|}{~~-- Grid Refiner} & \underline{75.36}$_{\pm0.27}$ & 63.41$_{\pm0.67}$ & 44.30$_{\pm0.51}$ & 42.44$_{\pm0.59}$ & \underline{48.04}$_{\pm1.07}$ \\
            
            \bottomrule
        \end{tabular}
    \caption{Overall F1-score (\%) on SciEvents. For $^\dagger$EAE-only models, trigger predictions are derived from Tagprime, which achieves the best ED performance among all baseline methods.}
    \label{tab:experiment_results_main}
    \end{threeparttable}
\end{table*}

\paragraph{Grid Refiner}
The initial grid representations encode pair-wise token relations independently.
To enable information propagation across related token pairs, we refine the grid with a stack of lightweight residual refinement blocks operating on the grid space.

Let $\mathbf{G}^{(0)}=\mathbf{G}$ and $\mathbf{G}^{(k)} \in \mathbb{R}^{l \times l \times C_g}$ denotes the grid features after the $k$-th refinement layer.
Each block updates the grid by aggregating information from local neighborhoods and applying a residual transformation:
\begin{equation}
\mathbf{G}^{(k+1)}
=
\text{Norm}\!\left(
\mathbf{G}^{(k)} + \mathcal{F}\!\left(\mathbf{G}^{(k)}\right)
\right),
\end{equation}
where $\mathcal{F}(\cdot)$ denotes a learnable local aggregation function on the grid, and is instantiated as stacked 2D convolutional refinement blocks in our implementation.
After $K$ refinement layers, we obtain the refined grid representations $\tilde{\mathbf{G}}=\mathbf{G}^{(K)}$.

\subsection{Training and Inference}
\label{sec:train_and_inference}
\paragraph{Loss Function}
Given $\tilde{\mathbf{G}} \in \mathbb{R}^{l \times l \times C_g}$, we project each grid cell to relation logits via a linear classifier, yielding
$\mathbf{Y} \in \mathbb{R}^{l \times l \times |R|}$.
Since multiple relation types may simultaneously hold for a token pair, we formulate grid prediction as a multi-label classification problem.

We adopt a multi-label categorical cross-entropy loss \cite{su2022zlprnovellossmultilabel}, which jointly optimizes positive and negative labels without requiring a predefined number of active labels.
Formally, for each grid cell $(i,j)$, the loss is defined as
\begin{equation}
\mathcal{L}_{i,j}
=
\log(1 + \sum_{r \in \Omega^{-}} e^{y_{i,j}^{r}})
+
\log(1 + \sum_{r \in \Omega^{+}} e^{-y_{i,j}^{r}}),
\end{equation}
where $\Omega^{+}$ and $\Omega^{-}$ denote the sets of positive and negative relation types for $(x_i, x_j)$, respectively.

\paragraph{Inference}
Following the multi-label formulation, we obtain the predicted relation set for each grid cell by a zero-threshold decision:
\begin{equation}
\hat{\mathbf{M}}_{i,j,r} = \mathbb{I}\left[y_{i,j}^{r} > 0\right],
\end{equation}
where $\hat{\mathbf{M}} \in \{0,1\}^{l \times l \times |R|}$ is the binary word-word relation grid.
We then decode $\hat{\mathbf{M}}$ into a set of events by reconstructing nuggets and linking arguments to triggers.
Specifically, we (1) recover nugget spans by traversing \textsc{HTL} and validating them with a closing \textsc{THL}-type, and (2) attach argument nuggets to trigger nuggets using \textsc{EAL} and schema constraints.
Appendix~\ref{sec:grid_decode} presents the detailed decoding algorithm.

\section{Experiment}
\subsection{Experiment Settings}
\paragraph{Evaluation Metrics}
Four standard metrics are adopted: trigger identification (TI), trigger classification (TC), argument identification (AI), and argument classification (AC).
In addition, we introduce event correlation (EC) to evaluate the extraction of hierarchical sub-event relations. 
Specifically, when the trigger of one event appears as an argument of another event, the two events are considered correlated through their triggers.
An evaluation example is provided in Appendix~\ref{sec:metrics_demonstration}.

\paragraph{Baselines}
We conduct a comprehensive evaluation of state-of-the-art and recent EE models, which can be broadly categorized into three groups:
(1) Global information extraction models that jointly model entities, relations, and events within a unified framework; %, including OneIE, UTC-IE and ScentedEAE ; 
(2) Discriminative EE models that formulate EE as token classification or sequential labeling problems; %, including: TagPrime, EEQA, PAIE and DEEIA.
(3) Generative EE models that generate extractions via question answering or through a well-designed generative schema. %, including: BartGen, DEGREE, AMPERE and KnowCoder.

For a fair comparison, when evaluating EAE-only methods, we first apply a best-performing ED method to extract event triggers. The EAE-only methods then perform argument extraction conditioned on these predicted triggers.

\paragraph{Implementations}
We randomly split SciEvents into training, development, and test sets with a ratio of 80\%/10\%/10\%. Each model is evaluated over three independent runs, and the average performance is reported. 
For models with the same architecture, we use the same pretrained backbone.
Additional details are provided in Appendix~\ref{sec:implementations}.

\begin{table*}[ht]    
    \small
    \centering
    \begin{threeparttable}
        \begin{tabular}{l|ccccccc}
            \toprule
            \multirow{2}{*}{\bf Model} & Discontinuous & \multicolumn{2}{|c|}{Overlapping} & Reverse-order & \multicolumn{3}{|c}{Sub-event} \\
            ~ & AC  & TC & AC & AC  & TC & AC & EC \\
            \midrule
            OneIE & -- & \underline{62.20}$_{\pm1.07}$ & 17.33$_{\pm0.98}$ & -- & 51.19$_{\pm0.69}$ & 37.41$_{\pm1.51}$ & 41.39$_{\pm2.72}$ \\
            ScentedEAE & -- & 55.77$_{\pm2.40}$ & 11.05$_{\pm1.23}$ & -- & 43.23$_{\pm2.35}$ & 38.00$_{\pm2.19}$ & 33.04$_{\pm1.45}$ \\
            \midrule
            Tagprime & -- & 55.03$_{\pm1.59}$ & 18.11$_{\pm1.13}$ & -- & 53.84$_{\pm0.81}$ & 47.89$_{\pm0.39}$ & 48.11$_{\pm2.45}$ \\
            EEQA & -- & 17.02$_{\pm0.80}$ & 2.33$_{\pm0.02}$ & -- & 53.50$_{\pm0.92}$ & 44.81$_{\pm1.35}$ & 45.16$_{\pm1.21}$ \\
            PAIE & -- & 49.62$_{\pm1.71}$ & 13.18$_{\pm0.67}$ & -- & 53.66$_{\pm1.75}$ & 47.34$_{\pm1.01}$ & 49.08$_{\pm1.75}$ \\
            DEEIA & -- & 51.31$_{\pm2.84}$ & 13.13$_{\pm1.21}$ & -- & 42.73$_{\pm1.20}$ & 32.80$_{\pm1.67}$ & 31.30$_{\pm3.73}$ \\
            \midrule
            % BartGen & 2.17$_{\pm0.00}$ & 32.53$_{\pm0.00}$ & 9.84$_{\pm0.00}$ & 0.00$_{\pm0.00}$ & 48.95$_{\pm0.00}$ & 40.00$_{\pm0.00}$ & 33.18$_{\pm0.00}$ \\
            BartGen & 2.74$_{\pm0.18}$ & 31.98$_{\pm1.29}$ & 10.58$_{\pm0.44}$ & 0.00$_{\pm0.00}$ & 52.25$_{\pm0.13}$ & 43.61$_{\pm0.31}$ & 40.19$_{\pm2.46}$ \\
            % DEGREE & 0.00$_{\pm0.00}$ & 8.01$_{\pm0.00}$ & 2.67$_{\pm0.00}$ & 0.00$_{\pm0.00}$ & 28.92$_{\pm0.00}$ & 20.68$_{\pm0.00}$ & 12.72$_{\pm0.00}$ \\
            DEGREE & 0.00$_{\pm0.00}$ & 24.93$_{\pm1.74}$ & 7.16$_{\pm0.54}$ & 0.00$_{\pm0.00}$ & 39.24$_{\pm0.51}$ & 30.16$_{\pm0.91}$ & 19.97$_{\pm1.23}$ \\
            % KnowCoder & 0.00$_{\pm0.00}$ & 18.56$_{\pm0.96}$ & 6.45$_{\pm0.33}$ & 0.00$_{\pm0.00}$ & 42.36$_{\pm1.32}$ & 34.81$_{\pm0.89}$ & 40.33$_{\pm2.41}$ \\
            KnowCoder & 0.00$_{\pm0.00}$ & 26.18$_{\pm1.12}$ & 6.93$_{\pm0.39}$ & 0.00$_{\pm0.00}$ & 42.36$_{\pm1.32}$ & 34.81$_{\pm0.89}$ & 40.33$_{\pm2.41}$ \\
            \midrule
            EXCEEDS & \underline{13.86}$_{\pm2.29}$ & \textbf{62.46}$_{\pm3.86}$ & \textbf{22.46}$_{\pm2.01}$ & 7.27$_{\pm5.14}$ & \underline{55.13}$_{\pm0.32}$ & \textbf{48.32}$_{\pm0.18}$ & \textbf{51.15}$_{\pm0.56}$ \\
            ~-- Contextual & 13.41$_{\pm1.34}$ & 59.72$_{\pm7.56}$ & 20.43$_{\pm3.20}$ & \underline{9.55}$_{\pm4.41}$ & 54.72$_{\pm1.13}$ & 47.64$_{\pm0.85}$ & \underline{50.18}$_{\pm1.30}$ \\
            ~-- Grid Refiner & \textbf{14.98}$_{\pm1.19}$ & 61.81$_{\pm2.86}$ & \underline{22.07}$_{\pm0.80}$ & \textbf{11.99}$_{\pm2.12}$ & \textbf{55.28}$_{\pm0.91}$ & \underline{48.04}$_{\pm1.07}$ & 49.11$_{\pm0.98}$ \\
            \bottomrule
        \end{tabular}
    \end{threeparttable}
    \caption{F1-score~(\%) on complex nuggets and events. -- indicates that the method cannot be evaluated.}
    \label{tab:experiment_results_complex_forms}
\end{table*}
\subsection{Experiment Results}

\paragraph{Overall Results}
Table~\ref{tab:experiment_results_main} presents the overall performance of different models on SciEvents. Results with precision and recall score can be found in Appendix~\ref{sec:full_overall_results}. Overall, EXCEEDS achieves the strongest performance on most evaluation metrics, particularly on AI, AC and EC.
Notably, EXCEEDS outperforms the second-best model by 0.51\% on AC and 0.53\% on EC, indicating its advantage in extracting arguments and capturing hierarchical sub-event relations in scientific documents.

Among baseline families, global models show competitive performance on TI and TC, which can be attributed to their use of entity information during training. 
Discriminative models generally perform better on EAE than global and generative models, while generative approaches exhibit larger performance variance across metrics.

Ablation study shows that removing the contextual modeling module results in the most pronounced performance drop on EAE, with AC decreasing by 1.06\%, indicating that grid-based modeling critically relies on high-quality contextual token representations. Excluding the grid refiner also leads to consistent degradation across metrics. These results suggest the effectiveness of contextual encoding and grid refinement in EXCEEDS.
Despite these improvements, the overall performance on AC remains modest, indicating that SciEvents remains a challenging EE dataset.

\paragraph{Complex Scenarios Results}
Table~\ref{tab:experiment_results_complex_forms} reports model performance on complex scenarios, where evaluation is conducted on subsets filtered by specific information forms. 
Overall, EXCEEDS consistently outperforms all baselines across complex scenarios, demonstrating its effectiveness in extracting complex nuggets and events. However, performance on discontinuous, overlapping, and reverse-order scenarios is substantially lower than the overall results, indicating that these forms remain particularly challenging for current EE models, even with specialized modeling designs.

Notably, generative models do not exhibit advantages under complex scenarios. In particular, they show pronounced performance degradation on overlapping nuggets, a trend also observed in EEQA, which adopts a generation-style formulation. This suggests that directly generating textual outputs is insufficient for accurately capturing complex nugget structures. %, especially when multiple event mentions overlap within dense scientific contexts.

\paragraph{Error Analysis}

\begin{figure}[htbp]
    \centering
    \begin{subfigure}{0.46\linewidth}
        \centering
        \includegraphics[width=\linewidth]{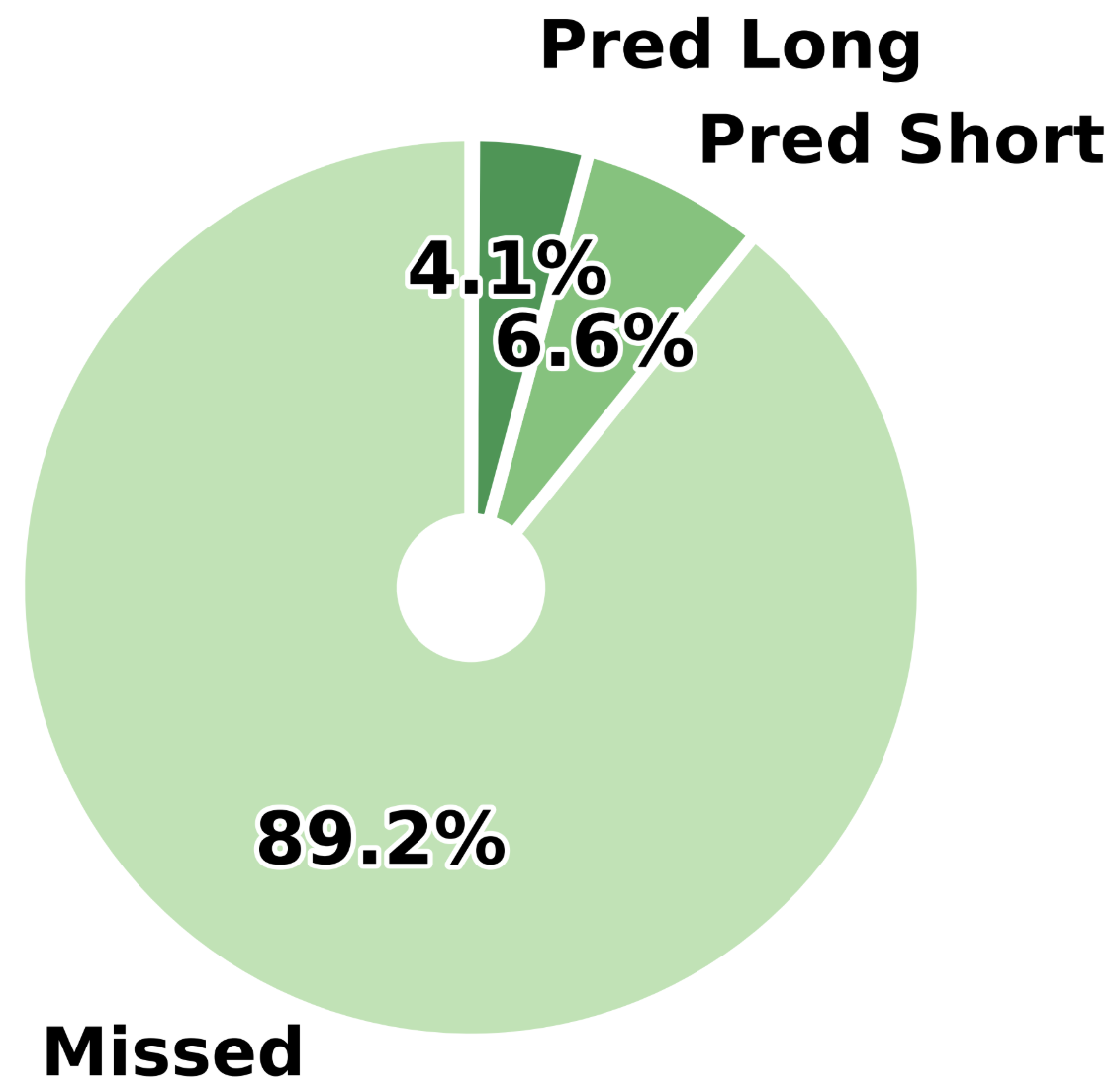}
        \caption{TI Error Distribution}
        \label{fig:ti_error_types_pie}
    \end{subfigure}\hfill
    \begin{subfigure}{0.53\linewidth}
        \centering
        \includegraphics[width=\linewidth]{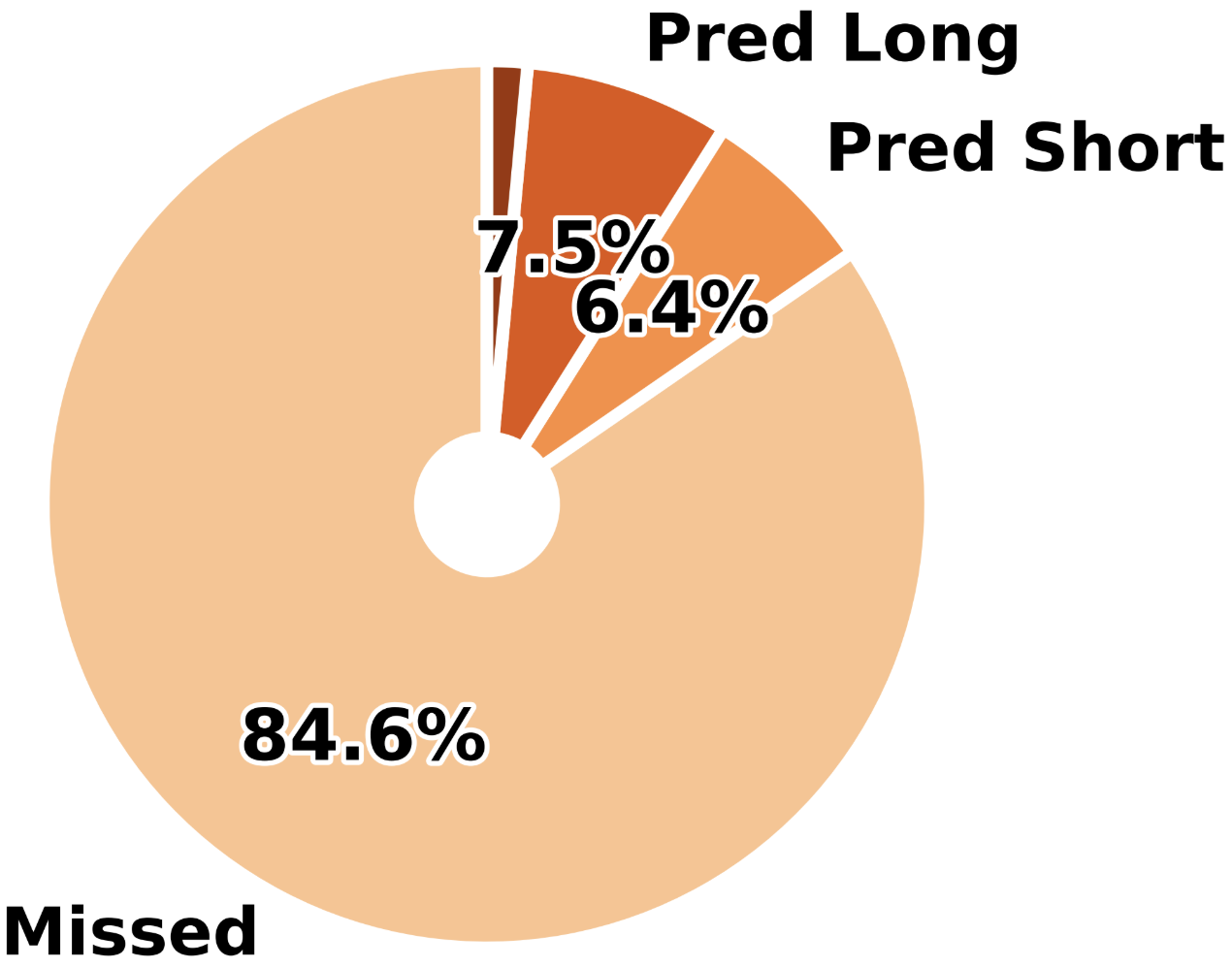}
        \caption{AI Error Distribution}
        \label{fig:ai_error_types_pie}
    \end{subfigure}

    \caption{Identification Error Distribution}
    \label{fig:identification_error_distribution}
\end{figure}

Figure~\ref{fig:identification_error_distribution} shows the identification error distributions for TI and AI, where errors are categorized into missed, predicted-long, predicted-short, and other overlap cases. 
Missed predictions dominate identification errors, accounting for 89.2\% of TI errors and 84.6\% of AI errors. 
This suggests that the primary challenge lies in failing to detect the presence of event mentions, rather than in inaccurately predicting their span boundaries. 
By contrast, boundary-related errors constitute a substantially smaller portion of the overall errors. This observation indicates nugget boundary imprecision is not the main bottleneck for identification performance. 
Overall, these results highlight that improving recall in dense scientific contexts remains a key challenge for event identification.

\begin{figure}[htbp]
    \centering
    \begin{subfigure}{\linewidth}
        \centering
        \includegraphics[width=\linewidth]{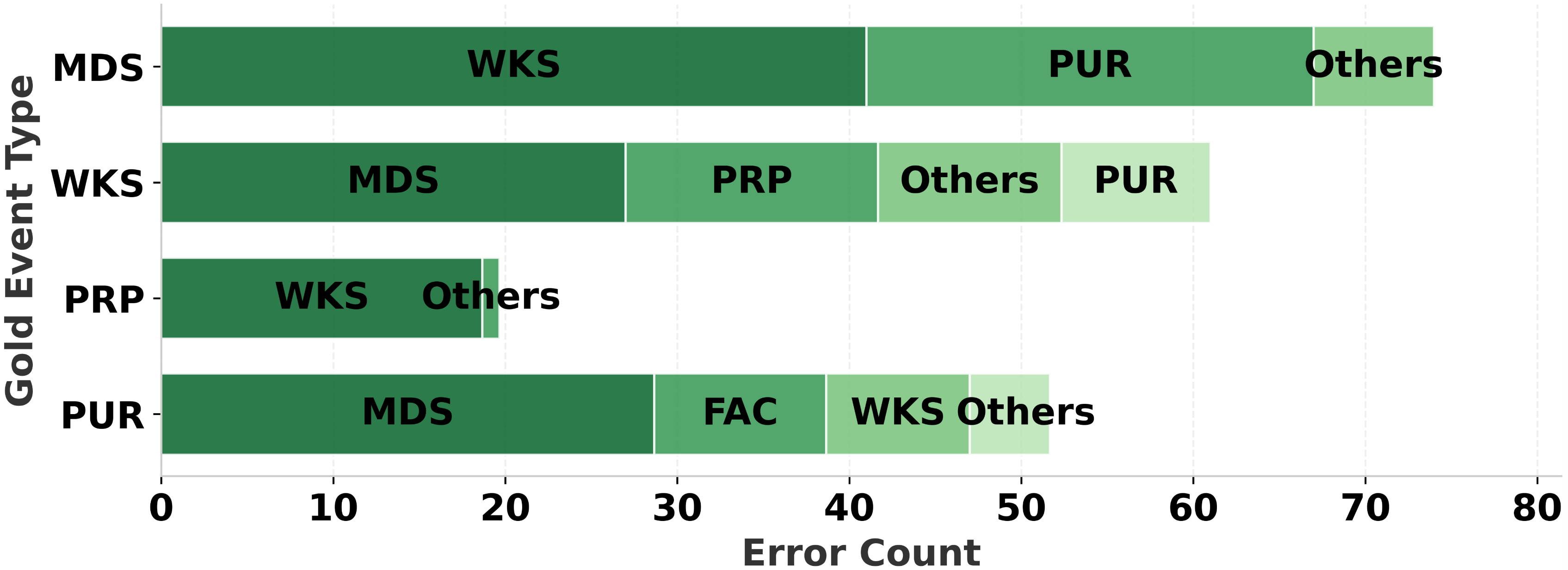}
        % \caption{TC Error Distribution}
        \label{fig:event_type_error_bar}
    \end{subfigure}

    \begin{subfigure}{\linewidth}
        \centering
        \includegraphics[width=\linewidth]{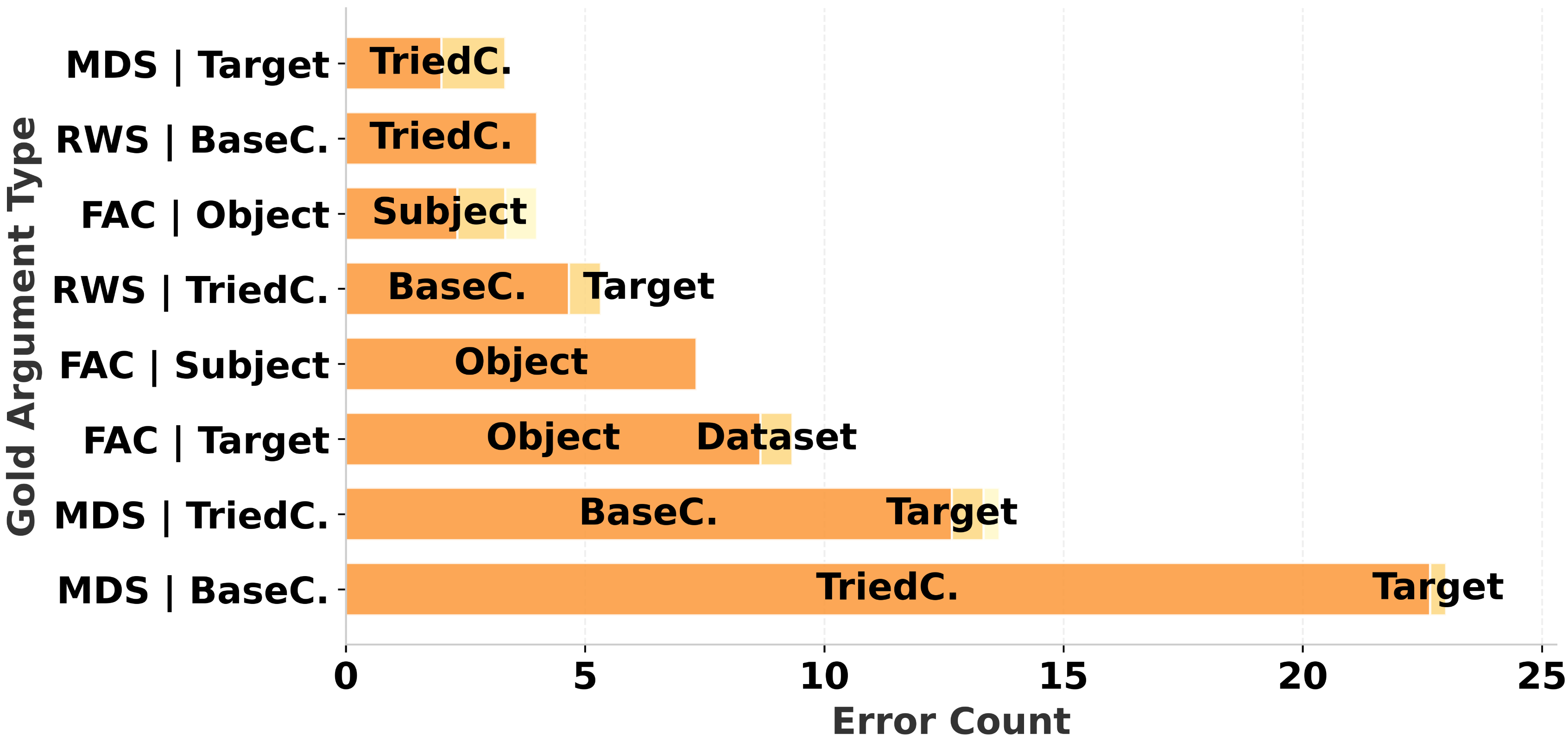}
        % \caption{AC Error Distribution}
        \label{fig:argument_type_error_bar}
    \end{subfigure}
    
    \caption{Classification Error Distribution. Y-axis denotes the misclassified gold type and each stacked bar indicates an incorrect predicted type and its frequency.}
    \label{fig:classification_error_distribution}
\end{figure}

Figure~\ref{fig:classification_error_distribution} presents the misclassification patterns in EXCEEDS for TC and AC. Overall, errors in both TC and AC are dominated by confusions between semantically similar types. 
For TC, frequent confusions occur between closely related event types, notably \textit{MDS} and \textit{WKS}. For AC, the most common errors also arise from fine-grained role distinctions, such as \textit{TriedC.} \textit{vs.} \textit{BaseC.} and \textit{Subject} \textit{vs.} \textit{Object}. 
These patterns suggest that improving classification on SciEvents likely requires more fine-grained modeling of subtle semantic distinctions, \textit{e.g.}, specialized disambiguation components, or incorporating schema/template-level cues \cite{ma-etal-2022-prompt, liu-etal-2024-beyond-single} to better differentiate similar types.

\section{Conclusion}
We address the challenge of understanding scientific domain with complex event structures through EE by introducing SciEvents, a large-scale document-level dataset with a refined schema tailored to the scientific domain, and EXCEEDS, a nugget-based grid modeling EE approach. Experiments show that SciEvents reflects the information-dense and structurally complex nature of scientific texts, while EXCEEDS achieves strong performance. Further analysis indicates that SciEvents remains a challenging dataset and suggests directions for future improvements.

\section*{Acknowledgments}
This work was supported by the National Key Research and Development Program of China (No. 2024YFF0908200).

\section*{Limitations}
\paragraph{Abstract-Level Data Scope} 
SciEvents is constructed from paper abstracts, rather than full-length scientific articles. While abstracts typically provide concise and information-dense summaries of scientific contributions, they do not capture all event mentions that may appear in the main body of a paper. In particular, important information conveyed through figures, tables, equations, or cross-section references is not covered in our current dataset. As a result, SciEvents does not account for multimodal or long-range contextual signals that may be essential for comprehensive scientific event understanding. Extending the dataset to full-text articles and incorporating multimodal information remains an important direction for future work.

\paragraph{Limited Domain Coverage}
SciEvents is constructed from ACL conference abstracts, which primarily represent the NLP sub-domain of scientific literature. As a result, the dataset does not cover the full diversity of writing styles, terminologies, and event structures present in other scientific disciplines.

This design choice is intended to provide a relatively controlled setting for studying dense and structurally complex event patterns, while reducing variability introduced by cross-domain differences. Nevertheless, the restricted domain coverage may limit the generalizability of models trained on SciEvents to broader scientific contexts. Extending the dataset to a wider range of scientific domains remains an important direction for future work.

% Bibliography entries for the entire Anthology, followed by custom entries
%\bibliography{anthology,custom}
% Custom bibliography entries only
\bibliography{main}

\appendix

\begin{algorithm}[htbp]
\small
\caption{Decoding grid into events}
\label{alg:decode}
\KwIn{Binary grid $\hat{\mathbf{M}} \in \{0,1\}^{l \times l \times |R|}$; label vocabulary $\mathcal{V}$ (including \textsc{HTL}, \textsc{EAL}, and \textsc{THL}-types); ontology checker $\textsc{Valid}(t_e, t_a)$.}
\KwOut{Event set $E$ in the target format.}

Initialize \texttt{Forward} $\leftarrow$ empty map \tcp*{\textsc{HTL}: head $\rightarrow$ next tokens}
Initialize \texttt{Tails} $\leftarrow$ empty map \tcp*{\textsc{THL}-type: head $\rightarrow$ possible tails}
Initialize \texttt{Links} $\leftarrow$ empty set \tcp*{\textsc{EAL}: (trigger-head, argument-head)}

\For{$i \leftarrow 1$ \KwTo $l$}{
  \For{$j \leftarrow 1$ \KwTo $l$}{
    $\mathcal{R}_{i,j} \leftarrow \{r \in R \mid \hat{\mathbf{M}}_{i,j,r}=1\}$\;
    \ForEach{$r \in \mathcal{R}_{i,j}$}{
      \uIf{$r = \textsc{HTL}$ \textbf{and} $i \neq j$}{
        \texttt{Forward}[$i$] $\leftarrow$ \texttt{Forward}[$i$] $\cup \{j\}$\;
      }
      \uElseIf{$r = \textsc{EAL}$}{
        \texttt{Links} $\leftarrow$ \texttt{Links} $\cup \{(i,j)\}$\;
      }
      \Else{
        \tcp{$r$ is a \textsc{THL}-type label indicating mention type}
        \texttt{Tails}[$j$] $\leftarrow$ \texttt{Tails}[$j$] $\cup \{i\}$\;
      }
    }
  }
}

\tcp{Step 1: recover nuggets by DFS over \textsc{HTL} and close with \textsc{THL}-type}
Initialize \texttt{Mentions} $\leftarrow \emptyset$\;
\ForEach{head $h$ in \texttt{Tails}}{
  Run DFS starting from $h$ following \texttt{Forward} edges to enumerate paths $p=[h,\dots,t]$\;
  Keep $p$ only if $t \in \texttt{Tails}[h]$ \tcp*{Heuristic (1): must be closed by \textsc{THL}-type}
  Add each kept path as a mention span into \texttt{Mentions}\;
}

\tcp{Step 2: assign mention types via closing \textsc{THL}-type and split triggers/arguments}
Initialize \texttt{Triggers} $\leftarrow \emptyset$, \texttt{Args} $\leftarrow \emptyset$\;
\ForEach{mention span $p=[h,\dots,t]$ in \texttt{Mentions}}{
  $\mathcal{T} \leftarrow \{r \in R \mid \hat{\mathbf{M}}_{t,h,r}=1 \ \wedge\ r \neq \textsc{HTL} \ \wedge\ r \neq \textsc{EAL}\}$\;
  \ForEach{$\tau \in \mathcal{T}$}{
    \uIf{$\tau \in T_E$}{\texttt{Triggers} $\leftarrow$ \texttt{Triggers} $\cup \{(p,\tau)\}$}
    \Else{\texttt{Args} $\leftarrow$ \texttt{Args} $\cup \{(p,\tau)\}$}
  }
}

\tcp{Step 3: build events by linking arguments to triggers via \textsc{EAL} + ontology constraints}
Initialize $E \leftarrow \{ e=(t_e, \text{trigger}=p, A=\emptyset) \mid (p,t_e)\in \texttt{Triggers}\}$\;
\ForEach{$(p_a, t_a) \in \texttt{Args}$}{
  Let $h_a$ be the head (first token) of $p_a$\;
  Find all triggers $(p_t, t_e)$ such that $(h_t, h_a)\in \texttt{Links}$ and $\textsc{Valid}(t_e,t_a)$\;
  \If{no such trigger exists}{
    \textbf{continue} \tcp*{Heuristic (2): drop arguments not attachable to any trigger}
  }
  \ForEach{matched trigger event $e$}{
    Add $(t_a, p_a)$ into $e.A$\;
  }
}
\Return{$E$}\;
\end{algorithm}

\section{Word-word Event Grid Decoding}
\label{sec:grid_decode}
To ensure the structural validity of decoded nuggets and events, we apply two pruning heuristics: 
(1) an \textsc{HTL} chain is kept only if it can be closed by a \textsc{THL}-type edge; (2) an argument nugget is kept only if it can be linked to at least one trigger nugget via \textsc{EAL} and passes the ontology constraint.

With these two pruning heuristics, Algorithm~\ref{alg:decode} summarizes the detailed decoding process.

Notably, to prevent potential excessively long decoding time during early training stages, we adopt a conservative training strategy. In early epochs, model predictions may be unstable and could assign a large number of labels to grid cells, which in the worst case may lead to an exponential number of candidate \textsc{HTL} chains during DFS-based decoding and significantly slow down validation. To avoid such degenerate cases, we skip validation in the initial training phase (typically the first few epochs) and enable regular validation after this phase.

\section{Evaluation Metrics Demonstration}
\label{sec:metrics_demonstration}
In SciEvents, there are 3 tasks and 5 kinds of metrics: (1) Trigger Extraction, also known as Event Detection, includes Trigger Identification (TI) and Trigger Classification (TC). (2) Event Argument Extraction includes Argument Identification (AI) and Argument Classification (AC). (3) Sub-Event Extraction includes Event Correlation (EC).

In SciEvents, a nugget serves as the basic unit for evaluation, and an exact-match criterion is applied when assessing nugget token spans.
Table~\ref{tab:prediction_example} shows an example with two prediction events. 
Table~\ref{tab:evaluation_example} shows how to evaluate these two prediction events.

\begin{table}[htbp]
    \small
    \centering
  \begin{tabular}{c|cc}
    \toprule   
    ~ & Event 1 & Event 2 \\
    \midrule 
    trigger & A (Type M) & C (Type N) \\
    \midrule
    \multirow{2}{*}{argument} & B (Type BT) & D (Type DT)\\
    ~ & C (Type CT) & ~\\
    \bottomrule
  \end{tabular}
  \caption{An example with two prediction events. A, B, C and D are token spans. M and N are event types. BT, CT and DT are argument types.}
  \label{tab:prediction_example}
\end{table}

\begin{table}[htbp]
    \small
    \centering
  \begin{tabular}{c|l}
    \toprule   
    \multirow{2}{*}{\textbf{Metric}} & What elements should \\
    ~ & match ground-truth \\
    \midrule 
    \multirow{2}{*}{TI} & A \\
    ~ & C \\
    \midrule
    \multirow{2}{*}{TC} & A+M \\
    ~ & C+N \\
    \midrule
    \multirow{3}{*}{AI} & A+M+B \\
    ~ & A+M+C \\
    ~ & C+N+D \\
    \midrule
    \multirow{3}{*}{AC} & A+M+B+BT \\
    ~ & A+M+C+CT \\
    ~ & C+N+D+DT \\
    \midrule
    EC & A+M+C+CT+N \\
    \bottomrule
  \end{tabular}
  \caption{An evaluation example for two prediction events. A, B, C and D are token spans. M and N are event types. BT, CT and DT are argument types.}
  \label{tab:evaluation_example}
\end{table}

\section{Implementation Details}
\label{sec:implementations}

\paragraph{Pretrained Backbone.} For all evaluated models, we adopt either BART-large \cite{lewis-etal-2020-bart} or RoBERTa-large \cite{liu2019robertarobustlyoptimizedbert} as the pretrained backbone to ensure a fair and controlled comparison across architectures. 
KnowCoder \cite{li-etal-2024-knowcoder}, which is based on large language models, employs LLaMA 2-7B \cite{touvron2023llama2openfoundation} as its pretrained backbone.

\paragraph{Hyperparameter Settings.} For EXCEEDS, the hyperparameter settings used in our implementation are reported in Table~\ref{tab:parameters}.
For the remaining models, we primarily apply a unified event extraction framework, TextEE \cite{huang-etal-2024-textee}, and adopt the hyperparameters recommended by TextEE. For models that are not supported by TextEE, we adapt SciEvents to their official code, and use the hyperparameter settings suggested in their implementations.

\begin{table}[htbp]
    \small
    \centering
  \begin{tabular}{l|c}
    \toprule   
    \textbf{Hyperparameter} & \textbf{Value} \\
    \midrule 
    Roberta-large Learning Rate & 1e-5 \\
    Warm Up Ratio & 0.1 \\
    Other Learning Rate & 1e-3 \\
    Batch Size & 2 \\
    Epoch & 20 \\
    Distance Embedding Size & 20 \\
    Bi-LSTM Hidden Size & 1024 \\
    Grid Channels ($C_g$) & 256 \\
    Grid Refiner Dropout Rate & 0.1 \\
    Other Dropout Rate & 0.5 \\
    Grid Refiner Layers ($K$) & 2 \\
    Grid Refiner Kernel & 3 \\
    \bottomrule
  \end{tabular}
  \caption{Hyperparameter used in EXCEEDS.}
  \label{tab:parameters}
\end{table}

\paragraph{Architecture-Specific Implementations.}
For global information extraction models, we leverage the nugget type information provided in SciEvents (described in Appendix~\ref{sec:nugget_type}) to facilitate entity training, following their original modeling assumptions.

For discriminative models, as most models rely on explicit start and end offsets for training and inference, discontinuous and reverse-order nuggets are not directly applicable under their modeling assumptions. Accordingly, these models are evaluated only on nugget instances with contiguous span representations.

%most approaches rely on nugget start and end offsets for training and inference. As a result, discontinuous nuggets and reverse-order nuggets are considered invalid inputs under these formulations. We therefore exclude these nugget types when preparing inputs for discriminative models and do not evaluate them on discontinuous or reverse-order complexity scenarios.

For generative models, input-output interfaces and evaluation scripts are modified to support raw text,  without relying on explicit span offsets, during training, inference, and evaluation. This adaptation ensures that generative approaches can be fairly evaluated on SciEvents under an offset-free setting.

\paragraph{Training and Inference Cost}
Most models are trained and evaluated on NVIDIA RTX 3090 GPUs. Experiments involving large language models are conducted on NVIDIA A800 80GB PCIe GPUs, where parameter-efficient fine-tuning with LoRA \cite{hu2021loralowrankadaptationlarge} is adopted. Table~\ref{tab:gpu_hour} reports the average training time \textbf{per epoch} and inference time for each model, measured in GPU hours, providing a reference for their computational cost. Appendix~\ref{sec:theoretical_analysis} provides a theoretical analysis of the computational complexity and memory overhead of EXCEEDS.

\begin{table}[htbp]
    \small
    \centering
    \begin{tabular}{c|l|cc}
        \toprule
        \multicolumn{2}{c}{\textbf{Model}} & Training & Inference \\
        \midrule
        \multirow{2}{*}{Global} & OneIE & 0.1816 & 0.0036\\
        ~ & ScentedEAE & 0.5249 & 0.0440 \\
        \midrule
        \multirow{8}{*}{ED \& EAE} & EEQA-ED & 0.0259 & 0.0056 \\
        ~ & EEQA-EAE & 5.2138 & 3.7292 \\
        ~ & Tagprime-ED & 0.2510 & 0.0014 \\
        ~ & Tagprime-EAE & 0.3910 & 0.0083 \\
        ~ & DEGREE & 0.3639 & 0.0067 \\
        ~ & KnowCoder-ED$^\dagger$ & 0.2381 & 0.2150 \\
        ~ & KnowCoder-EAE & 1.5869 & 0.2897 \\
        ~ & EXCEEDS (Ours) & 0.0609 & 0.1692 \\
        \midrule
        \multirow{3}{*}{EAE-only} & PAIE & 0.7536 & 0.0033 \\
        ~ & DEEIA & 0.1237 & 0.0165\\
        ~ & BartGen & 0.2950 & 0.0236 \\
        \bottomrule
    \end{tabular}
    \caption{Average GPU hours required per training epoch and for inference across different models. $^\dagger$is based on a large language model and is fine-tuned using parameter-efficient adaptation.}
    \label{tab:gpu_hour}
\end{table}

\section{Computational Complexity and Memory of EXCEEDS}
\label{sec:theoretical_analysis}
Let $l$ denote the input sequence length, $d$ the encoder hidden size, $C_g$ the grid channel size, $K$ the number of grid refinement layers, and $|R|$ the number of relation types of the word-word event grid. EXCEEDS consists of:

\begin{itemize}
    \item \textit{Contextual token encoding}: a pretrained transformer encoder (RoBERTa), followed by a BiLSTM and CLN. The transformer encoding has the standard complexity $O(l^2 d)$, while BiLSTM and CLN are $O(ld^2)$ and $O(ld)$, respectively.
    \item \textit{Pair-wise grid construction}: explicit construction of an $l \times l$ token-pair grid, followed by applying an MLP to each token pair to obtain grid features. This step scales as $O(l^2)$, with constants determined by the MLP width and feature dimensions.
    \item \textit{Grid refinement and classification}: application of $K$ lightweight 2D convolutional refinement blocks on the $l \times l$ grid, scaling as $O(Kl^2)$, with constants determined by kernel size and channel width. The final classifier head projects each grid cell from $C_g$ to $|R|$ relation logits, costing $O(l^2 C_g |R|)$ for a linear head.
\end{itemize}

Overall, the complexity of EXCEEDS is dominated by $O(l^2)$ terms. The memory footprint is dominated by storing the grid features and logits, \textit{i.e.}, $O(l^2 C_g + l^2 |R|)$, in addition to encoder activations.

\section{Full Experiment Results}
\label{sec:full_overall_results}

Table~\ref{tab:prf_trigger}, \ref{tab:prf_argument} and \ref{tab:prf_event} presents the full experiment results, including precision and recall scores.

\section{Schema of SciEvents}
\label{sec:schema}
The schema of SciEvents consists of nugget types and event types. In this section, we will introduce the description of each nugget type and the template of each event type.
\subsection{Nugget Types}
\label{sec:nugget_type}
There are 10 nugget types in SciEvents as follows:

\paragraph{Research Organization / Group (OG)} refers to a research team composed of people. Typical examples include: \textit{We}; \textit{Li et al., 2013}; \textit{They}.

\paragraph{Approach (APP)} refers to nouns, pronouns, and corresponding phrases that denote a complete method or algorithm with concrete inputs and outputs.
Typical examples include: \textit{... work}; \textit{... model}; \textit{... method}; \textit{... framework}; \textit{... network}; \textit{... algorithm}; \textit{baselines}; \textit{state-of-the-art}.

\paragraph{Module (MOD)} refers to nouns, pronouns, and corresponding phrases that denote components of a method or architecture, such as modules or algorithmic elements. A single MOD item is usually not detailed enough to constitute a full APP. Typical examples include: \textit{... encoders}; \textit{... decoders}; \textit{... module}; \textit{... process}; \textit{message propagation process}; \textit{beam search}.

\paragraph{Feature (FEA)} refers to nouns, pronouns, and corresponding phrases that denote features. Typical examples include: \textit{... information}; \textit{the first-order adjacency information}; \textit{the relationships between labeled edges}.

Note the difference among APP, MOD, and FEA: \textbf{APP} refers to a complete method with concrete inputs (\textit{e.g.}, a task) and outputs (\textit{e.g.}, the desired results of the task). \textbf{MOD} refers to a sub-process or component within the overall APP framework, such as a module or algorithmic element. \textbf{FEA} refers to features utilized during the execution of an APP or a MOD, such as \textit{positional information}, \textit{vector representations of part-of-speech tags}, or \textit{sentence length}.

\paragraph{Task (TAK)} refers to phrases that denote the intention or objective of a task optimization, \textit{i.e.}, the research focus or target point. These expressions are neutral. Typical examples include: \textit{graph-to-sequence modeling}; \textit{performance unimodal}; \textit{performance multimodal}; \textit{accuracy}; \textit{F1 score}; \textit{robustness}; \textit{reproducibility}; \textit{zero-shot translation quality}.

\paragraph{Dataset (DST)} refers to nouns, pronouns, and corresponding phrases that denote datasets used or relied upon when describing an artifact, research objective, or experimental conclusion. Typical examples include: \textit{TAC-KBP 2017 datasets}; \textit{Chinese multimodal NER dataset}; \textit{CNERTA}; \textit{training data}.

\paragraph{Limit (LIM)} refers to phrases that denote conditional or environmental limitations, often introduced with prepositions. Typical examples include: \textit{for a small number of confusing type pairs}; \textit{in existing verb metaphor detection benchmarks}; \textit{of the dynamic self-attention}.

\paragraph{Strength (STR)} refers to phrases that describe the advantages or strengths of an artifact, often with an evaluative or positive connotation. Typical examples include: \textit{state-of-the-art performance}.

\paragraph{Weakness (WEA)} refers to phrases that describe the disadvantages, shortcomings, or weaknesses of an artifact, often with an evaluative or negative connotation. Typical examples include: \textit{most of the mislabeling}; \textit{biases and failure cases of beam search}.

\paragraph{Degree (DEG)} refers to adjectives, adverbs, numerals, or other expressions that describe the degree or quantity of an event. Typical examples include: \textit{only}; \textit{not fully}; \textit{1.5\%}.

\subsection{Event Types}
There are 10 event types classified by four different rhetorical components as follows:

(1) \textit{General}. \textit{General} events occur in all four components.

\begin{table}[ht]
    \small
    \centering
    \begin{tabular}{c|c}
        \toprule   
        {\bf Argument Type} & Constrained Types \\
        \midrule 
    
        \multirow{2}{*}{Aim} & APP / MOD / FEA / DST /  \\
        ~ & STR / WEA / TAK \\
        Condition     & LIM \\
        Dataset       & DST \\
        \bottomrule
    \end{tabular}
    \caption{Schema of Purpose (PUR) Event}
    \label{tab:event_type_pur}
\end{table}
\paragraph{Purpose (PUR).} As the schema shown in Table~\ref{tab:event_type_pur}, the Purpose event describes: \textit{In order to deal with <Aim:arg1> under <Condition:arg2> circumstance on <Dataset:arg3> datasets.}

(2) \textit{Background} includes one kind of event type:

\begin{table}[ht]
    \small
    \centering
    \begin{tabular}{c|c}
        \toprule   
        {\bf Argument Type} & Constrained Types \\
        \midrule 
    
        \multirow{2}{*}{Target} & APP / MOD / FEA / DST /  \\
        ~ & STR / WEA / TAK \\
        Condition     & LIM   \\
        Dataset       & DST   \\
        \bottomrule
    \end{tabular}
    \caption{Schema of IntroduceTarget (ITT) Event}
    \label{tab:event_type_itt}
\end{table}

\paragraph{IntroduceTarget (ITT).} As the schema shown in Table~\ref{tab:event_type_itt}, the IntroduceTarget event describes: \textit{<Target:arg1> is the abstract research target under <Condition:arg2> circumstance on <Dataset:arg3> datasets in this paper.}

(3) \textit{Related Work} includes two kinds of event type:

\begin{table}[ht]
    \small
    \centering
    \begin{tabular}{c|c}
        \toprule   
        {\bf Argument Type} & Constrained Types  \\
        \midrule 
    
        Subject        & APP / MOD / FEA / DST                     \\
        BaseComponent  & APP / MOD / FEA / DST                     \\
        TriedComponent & APP / MOD / FEA / DST                     \\
        Condition      & LIM / E-RWS                               \\
        Dataset        & DST                                       \\
        \multirow{2}{*}{Target} & E-PUR / TAK / STR / WEA / \\
        ~ & APP / FEA / MOD \\
        \bottomrule
    \end{tabular}
    \caption{Schema of RelatedWorkStep (RWS) Event}
    \label{tab:event_type_rws}
\end{table}

\paragraph{RelatedWorkStep (RWS).} As the schema shown in Table~\ref{tab:event_type_rws}, the RelatedWorkStep event describes: \textit{Previously <Subject:arg1> on <Target:arg2> are mostly based on <BaseComponent:arg3> with <TriedComponent:arg4> under <Condition:arg5> circumstance on <Dataset:arg6> datasets.}

\begin{table}[ht]
    \small
    \centering
    \begin{tabular}{c|c}
        \toprule   
        {\bf Argument Type} & Constrained Types  \\
        \midrule 
        \multirow{2}{*}{Concern}       & APP / FEA / STR / WEA / \\
        ~ & MOD / DST \\
        \multirow{2}{*}{Fault}         & APP / FEA / STR / WEA / \\
        ~ & MOD / DST  \\ 
        Condition     & LIM / E-RWF / E-RWS               \\
        Dataset       & DST                               \\
        Target        & E-PUR / TAK / STR / WEA           \\
        Extent        & DEG                               \\
        \bottomrule
    \end{tabular}
    \caption{Schema of RelatedWorkFault (RWF) Event}
    \label{tab:event_type_rwf}
\end{table}
\paragraph{RelatedWorkFault (RWF).} As the schema shown in Table~\ref{tab:event_type_rwf}, the RelatedWorkFault event describes: \textit{Aiming to <Target:arg1>, to <Extent:arg5> degree, <Concern:arg2> has some <Fault:arg6> faults under <Condition:arg3> circumstance on <Dataset:arg4> datasets.}

(4) \textit{Methodology} includes three kinds of event type:

\begin{table}[ht]
    \small
    \centering
    \begin{tabular}{c|c}
        \toprule   
        {\bf Argument Type} & Constrained Types  \\
        \midrule 
        Proposer      & OG                          \\
        Content       & APP / FEA / MOD / DST / TAK \\
        Target        & E-PUR / TAK / FEA / WEA     \\
        \bottomrule
    \end{tabular}
    \caption{Schema of Propose (PRP) Event}
    \label{tab:event_type_prp}
\end{table}

\paragraph{Propose (PRP).} As the schema shown in Table~\ref{tab:event_type_prp}, the Propose event describes: \textit{In this paper, <Proposer:arg1> propose <Content:arg2> for <Target:arg3>.}

\begin{table}[ht]
    \small
    \centering
    \begin{tabular}{c|c}
        \toprule   
        {\bf Argument Type} & Constrained Types  \\
        \midrule 
    
        Researcher    & OG                                        \\
        \multirow{2}{*}{Content}       & APP / MOD / FEA / DST /   \\
        ~ & STR / WEA / TAK  \\
        Condition     & LIM                                       \\
        Dataset       & DST                                       \\
        \multirow{2}{*}{Target}        & E-PUR / TAK / STR / WEA /  \\
        ~ & APP / FEA / MOD \\
        \bottomrule
    \end{tabular}
    \caption{Schema of WorkStatement (WKS) Event}
    \label{tab:event_type_wks}
\end{table}

\paragraph{WorkStatement (WKS).} As the schema shown in Table~\ref{tab:event_type_wks}, the WorkStatement event describes: \textit{<Researcher:arg1> report <Content:arg2> under <Condition:arg3> circumstance on <Dataset:arg4> datasets for <Target:arg5>.}

\begin{table}[ht]
    \small
    \centering
    \begin{tabular}{c|c}
        \toprule   
        {\bf Argument Type} & Constrained Types  \\
        \midrule 
        BaseComponent  & APP / MOD / FEA / DST                     \\
        TriedComponent & APP / MOD / FEA / DST                     \\
        Condition      & LIM / E-MDS                               \\
        Dataset        & DST                                       \\
        \multirow{2}{*}{Target}   & E-PUR / TAK / STR / WEA / \\
        ~ & APP / FEA / MOD \\
        \bottomrule
    \end{tabular}
    \caption{Schema of MethodStep (MDS) Event}
    \label{tab:event_type_mds}
\end{table}

\paragraph{MethodStep (MDS).} As the schema shown in Table~\ref{tab:event_type_mds}, the MethodStep event describes: \textit{Our approach adopt <BaseComponent:arg1> with <TriedComponent:arg2> under <Condition:arg3> circumstance on <Dataset:arg4> datasets for <Target:arg5>.}

(5) \textit{Results} includes three kinds of event type:

\begin{table}[ht]
    \small
    \centering
    \begin{tabular}{c|c}
        \toprule   
        {\bf Argument Type} & Constrained Types  \\
        \midrule 
        Finder        & OG            \\
        Content       & E-FAC / E-CMP \\
        \bottomrule
    \end{tabular}
    \caption{Schema of Finding (FIN) Event}
    \label{tab:event_type_fin}
\end{table}

\paragraph{Finding (FIN).} As the schema shown in Table~\ref{tab:event_type_fin}, the Finding event describes: \textit{In experiments, <Finder:arg1> find or demostrate findings that <Content:arg2>.}

\begin{table}[ht]
    \small
    \centering
    \begin{tabular}{c|c}
        \toprule   
        {\bf Argument Type} & Constrained Types  \\
        \midrule 
        Arg1          & E-FAC / APP / MOD / FEA / DST \\
        Arg2          & E-FAC / APP / MOD / FEA / DST \\
        Condition     & LIM / E-FAC                   \\
        Dataset       & DST                           \\
        Result        & STR / WEA                     \\
        Metrics       & TAK                           \\
        Extent        & DEG                           \\
        \bottomrule
    \end{tabular}
    \caption{Schema of ExperimentCompare (CMP) Event}
    \label{tab:event_type_cmp}
\end{table}

\paragraph{ExperimentCompare (CMP).} As the schema shown in Table~\ref{tab:event_type_cmp}, the ExperimentCompare event describes: \textit{Experimental results show that the <Metrics:arg6> of <Arg1:arg1> is <Extent:arg2> <Result:arg3> than <Arg2:arg4> under <Condition:arg5> circumstance on <Dataset:arg7> datasets.}

\begin{table}[ht]
    \small
    \centering
    \begin{tabular}{c|c}
        \toprule   
        {\bf Argument Type} & Constrained Types  \\
        \midrule 
        \multirow{2}{*}{Subject}       & APP / MOD / FEA / STR /  \\
        ~ & WEA / TAK / DST \\
        \multirow{2}{*}{Object}        & APP / MOD / FEA / STR /  \\
        ~ & WEA / TAK / DST \\
        Condition     & LIM / E-FAC                             \\
        Reason        & LIM / E-FAC                             \\
        Dataset       & DST                                     \\
        Target        & E-PUR / TAK / STR / WEA                 \\
        Extent        & DEG                                     \\
        \bottomrule
    \end{tabular}
    \caption{Schema of OutcomeFact (FAC) Event}
    \label{tab:event_type_fac}
\end{table}
\paragraph{OutcomeFact (FAC).} As the schema shown in Table~\ref{tab:event_type_fac}, the OutcomeFact event describes: \textit{Experimental results show that <Subject:arg1> can <Extent:arg2> provide <Object:arg3> for <Target:arg4> under <Condition:arg5> circumstance on <Dataset:arg6> datasets because <Reason:arg7> reasons.}

\section{Example of Document-Level Event Annotation}
\label{sec:annotaion_example}
Table~\ref{tab:case_f63de5c2} presents a fully annotated example document from SciEvents, illustrating all event instances annotated within a single document. 
For each event, we show the corresponding trigger nugget, argument nuggets, and their semantic roles, as well as hierarchical sub-event relations when applicable. 
This example is provided to demonstrate the density and structural complexity of event annotations in scientific documents, and to facilitate a clearer understanding of the annotation schema and evaluation setup.

\section{Distributions in SciEvents}
\label{sec:scievents_distribution}
In this section, we will present comprehensive distributions in SciEvents.

\begin{figure*}[ht]
  \centering
  \includegraphics[width=\linewidth]{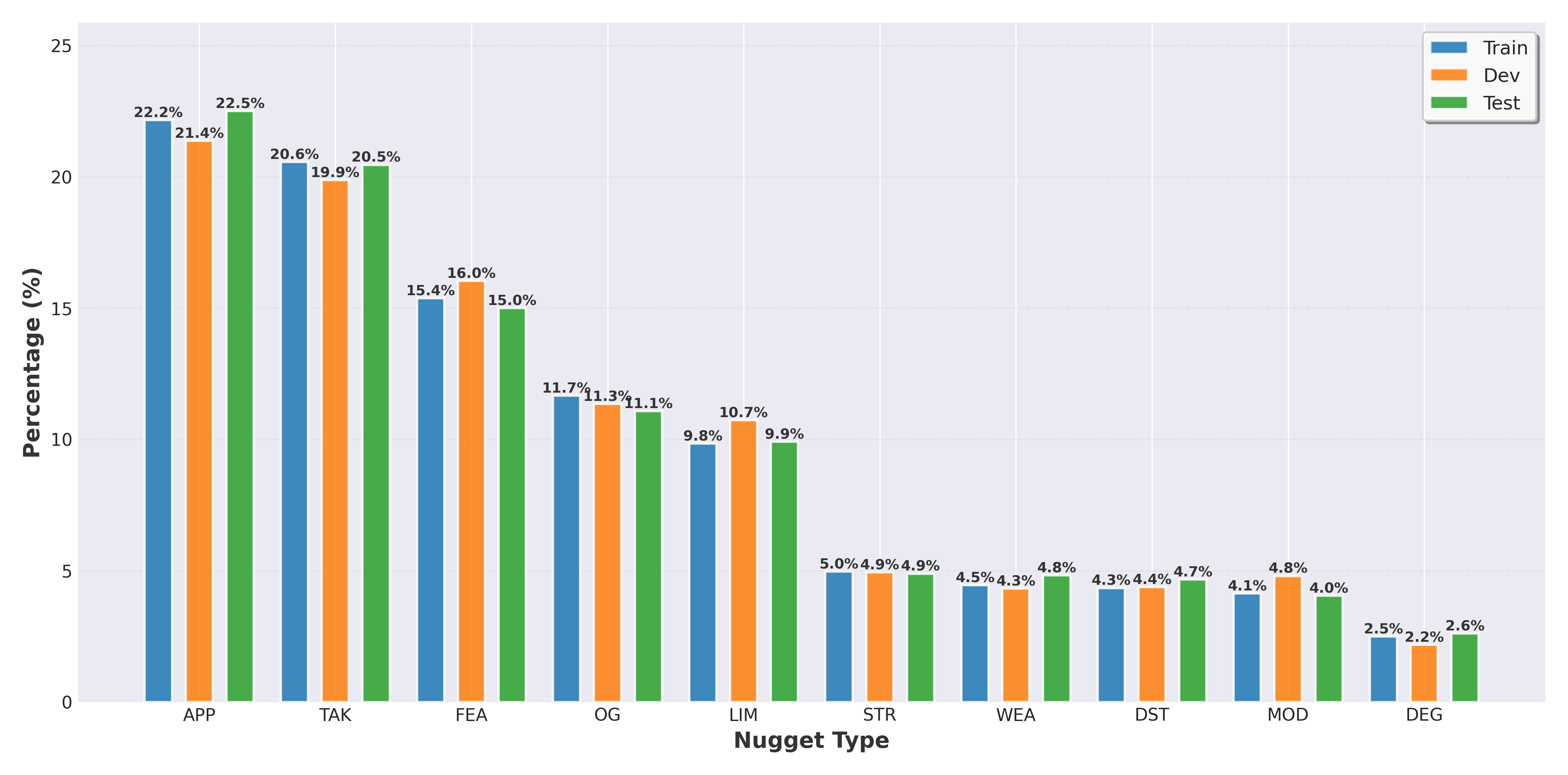}
  \caption{Percentage Distribution of Nugget Types across Train, Develop and Test Splits}
  \label{fig:nugget_percentage_distribution}
\end{figure*}

\paragraph{Nugget Type Distribution.} Table~\ref{tab:nugget_distribution} and Figure~\ref{fig:nugget_percentage_distribution} present distribution of nugget types across train, develop and test splits.

\begin{figure*}[ht]
  \centering
  \includegraphics[width=\linewidth]{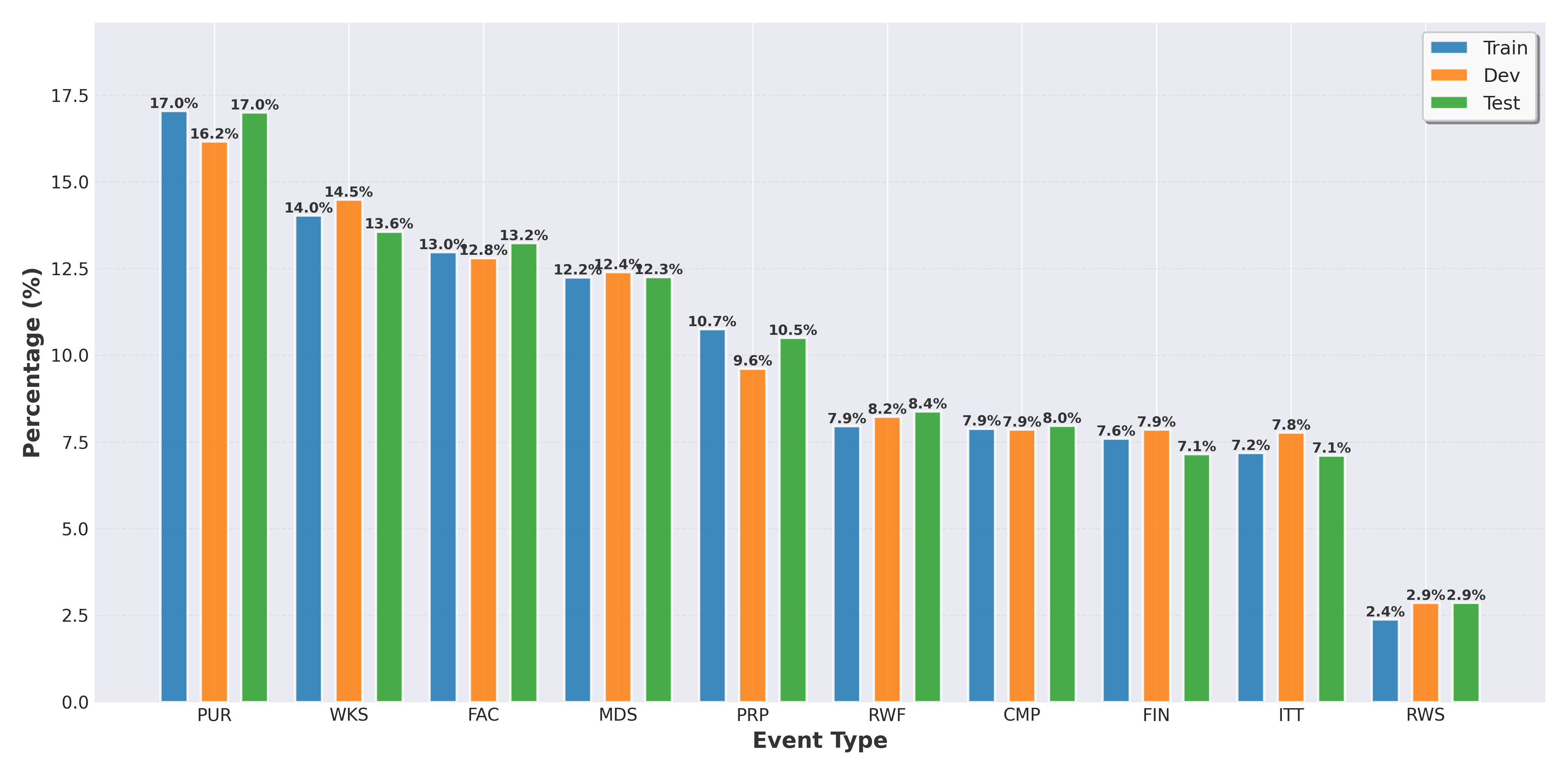}
  \caption{Percentage Distribution of Event Types across Train, Develop and Test Splits}
  \label{fig:event_type_percentage_distribution}
\end{figure*}

\paragraph{Event Type Distribution.} Table~\ref{tab:event_distribution} and Figure~\ref{fig:event_type_percentage_distribution} present distribution of event types across train, develop and test splits.

\begin{figure*}[ht]
  \centering
  \includegraphics[width=\linewidth]{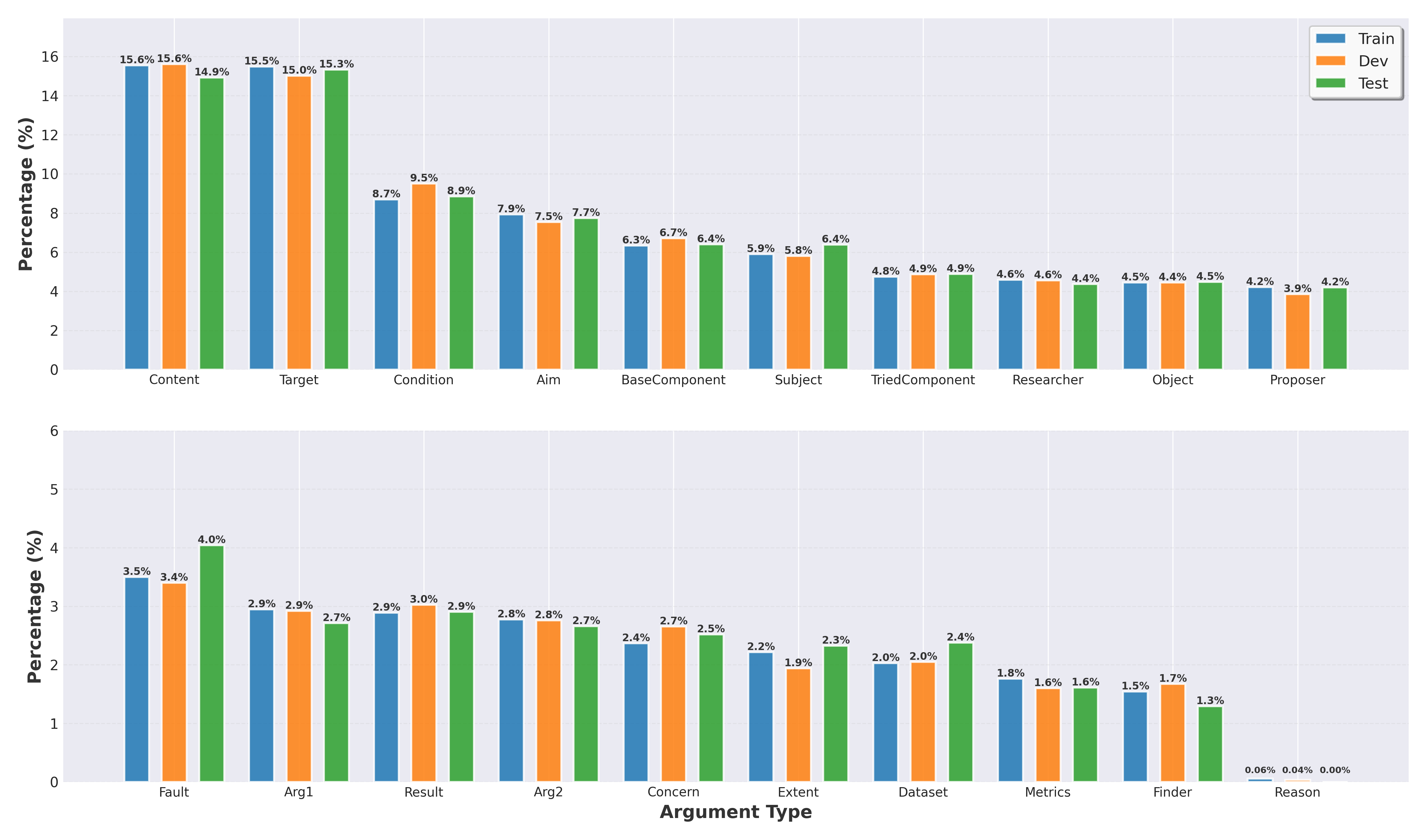}
  \caption{Percentage Distribution of Argument Types across Train, Develop and Test Splits}
  \label{fig:argument_type_percentage_distribution}
\end{figure*}

\paragraph{Argument Type Distribution.} Table~\ref{tab:argument_distribution} and Figure~\ref{fig:argument_type_percentage_distribution} present distribution of argument types across train, develop and test splits.

\begin{figure*}[ht]
  \centering
  \includegraphics[width=\linewidth]{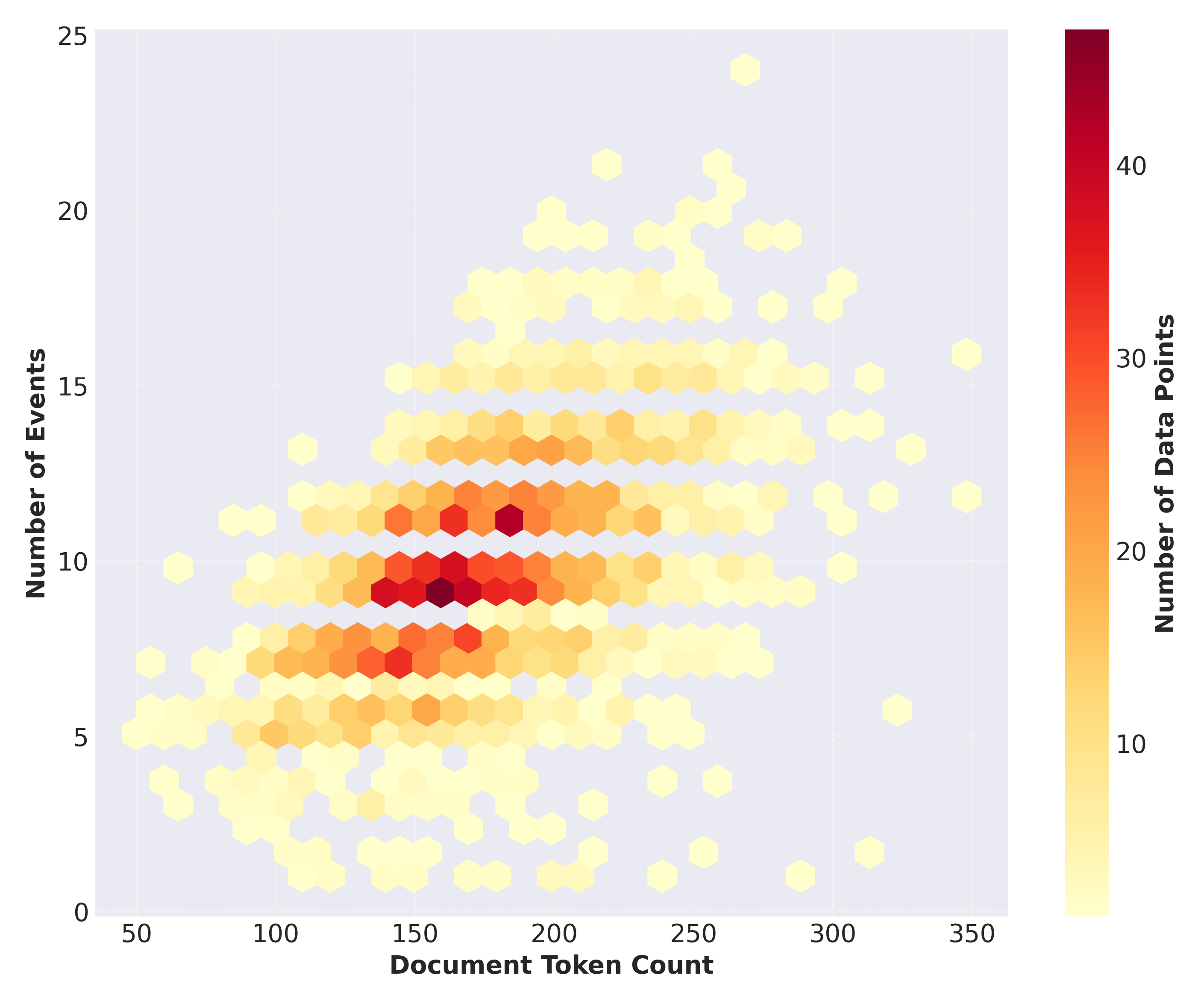}
  \caption{Document Length-Event Instance Distribution}
  \label{fig:document_length_event_instance_distribution}
\end{figure*}

\paragraph{Document Length-Event Instance Distribution.} Figure~\ref{fig:document_length_event_instance_distribution} present the distribution of document length versus event instance.

\begin{figure*}[t]
    \centering

    % ---------- Row 1 ----------
    \begin{subfigure}{0.32\linewidth}
        \centering
        \includegraphics[width=\linewidth]{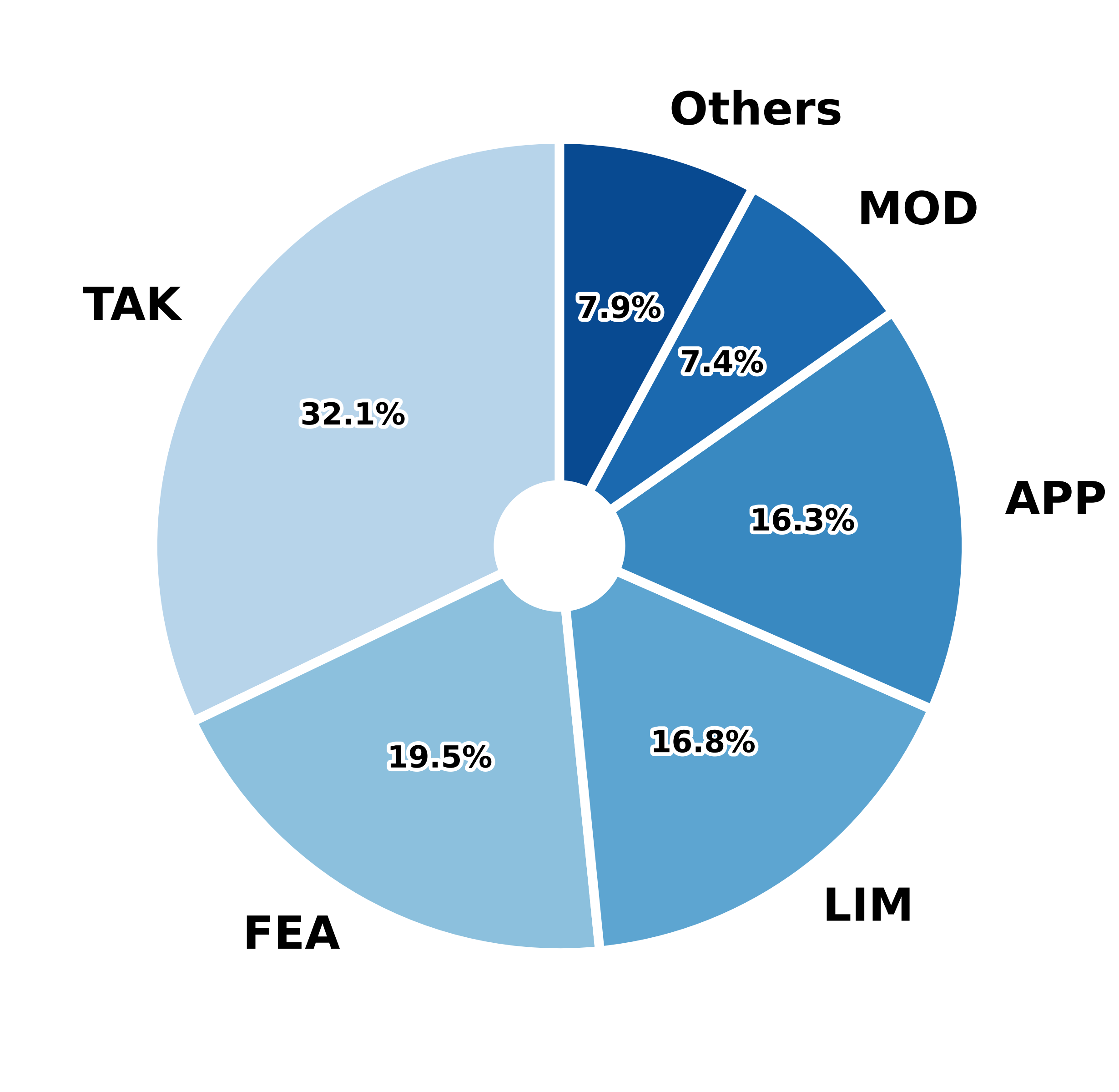}
        \caption{Discontinuous Nugget: Nugget Type}
        \label{fig:discontinuous_nugget_type}
    \end{subfigure}\hfill
    \begin{subfigure}{0.32\linewidth}
        \centering
        \includegraphics[width=\linewidth]{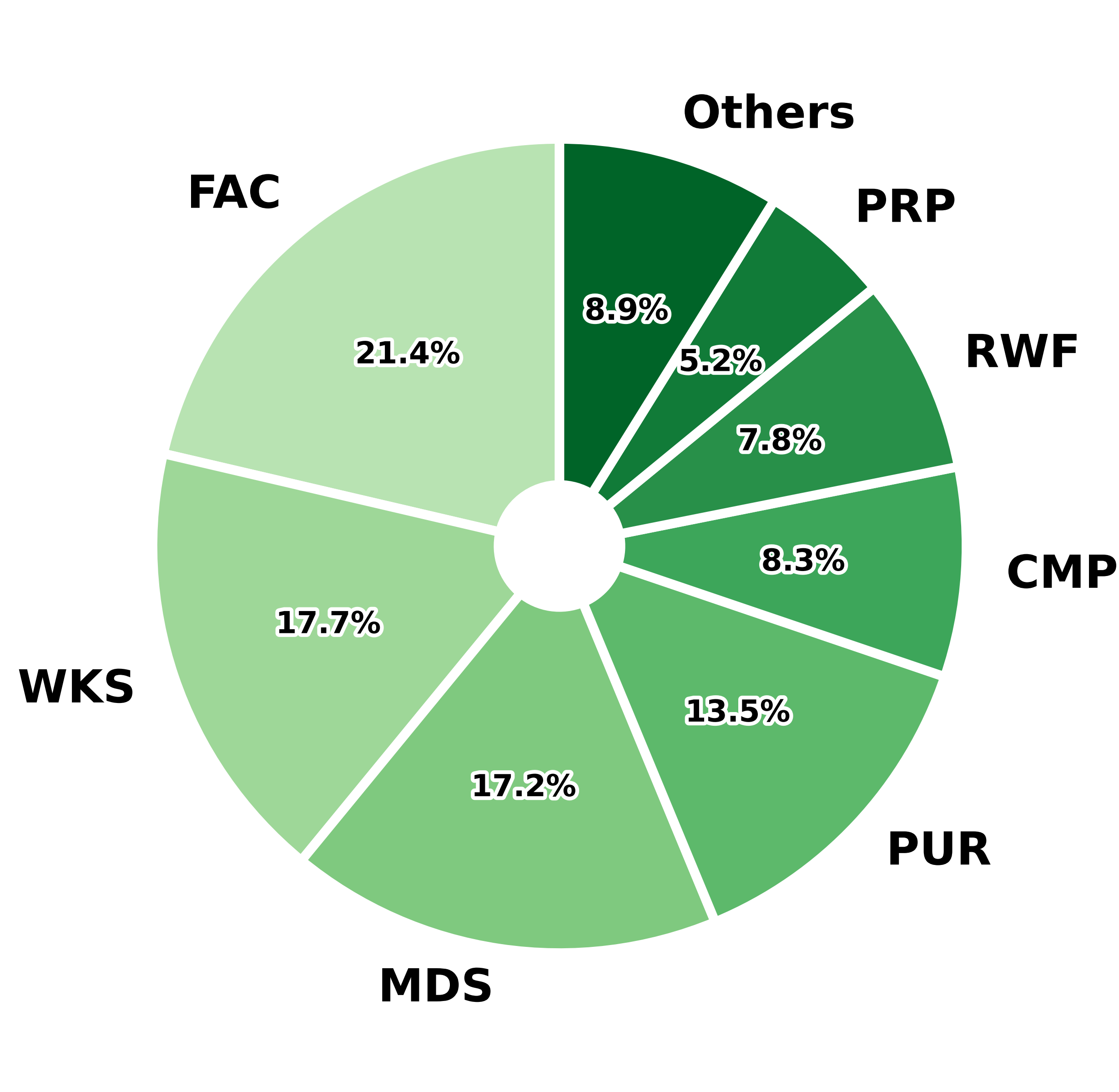}
        \caption{Discontinuous Nugget: Event Type}
        \label{fig:discontinuous_event_type}
    \end{subfigure}\hfill
    \begin{subfigure}{0.32\linewidth}
        \centering
        \includegraphics[width=\linewidth]{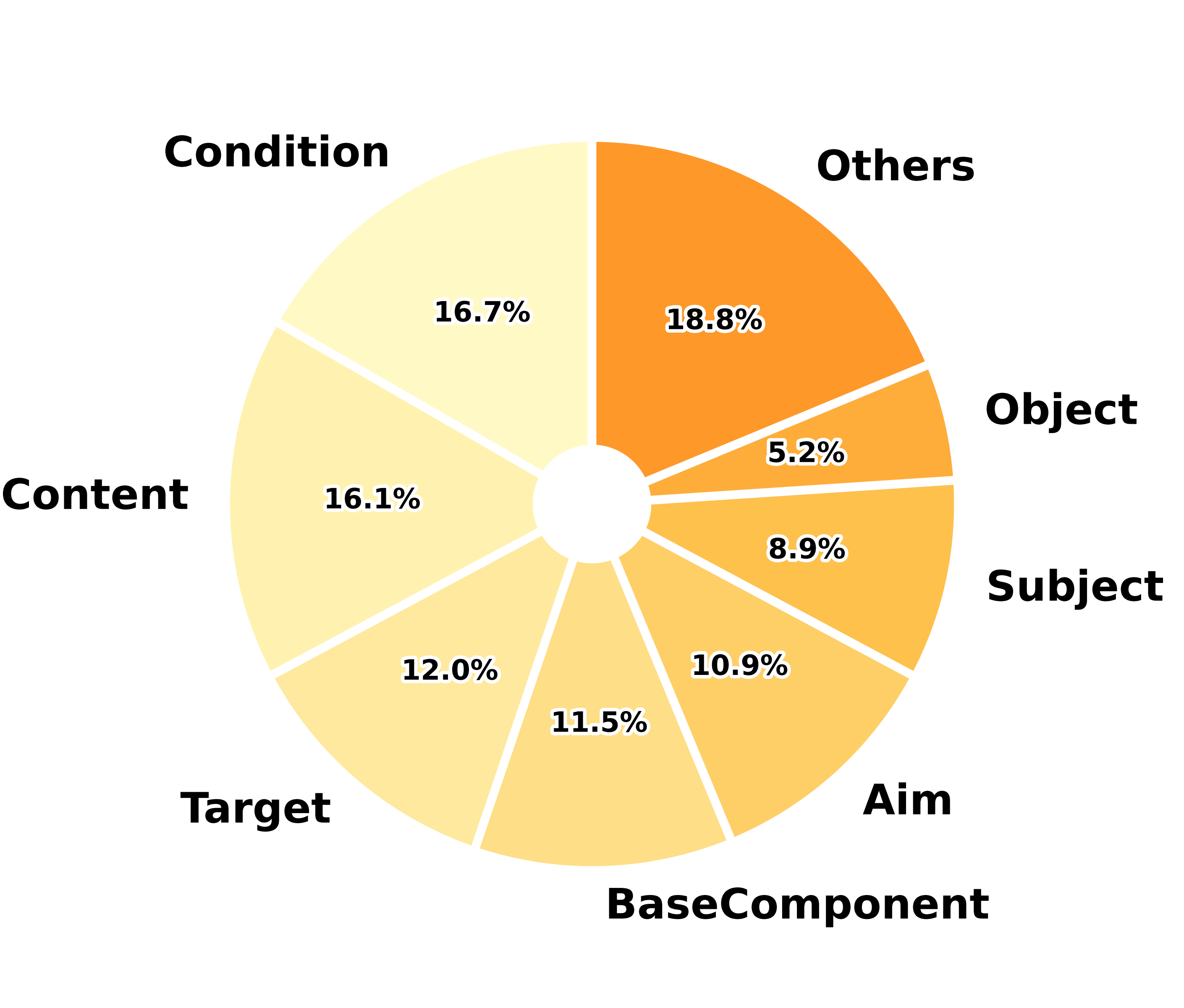}
        \caption{Discontinuous Nugget: Arg Type}
        \label{fig:discontinuous_argument_type}
    \end{subfigure}

    % ---------- Row 2 ----------
    \begin{subfigure}{0.32\linewidth}
        \centering
        \includegraphics[width=\linewidth]{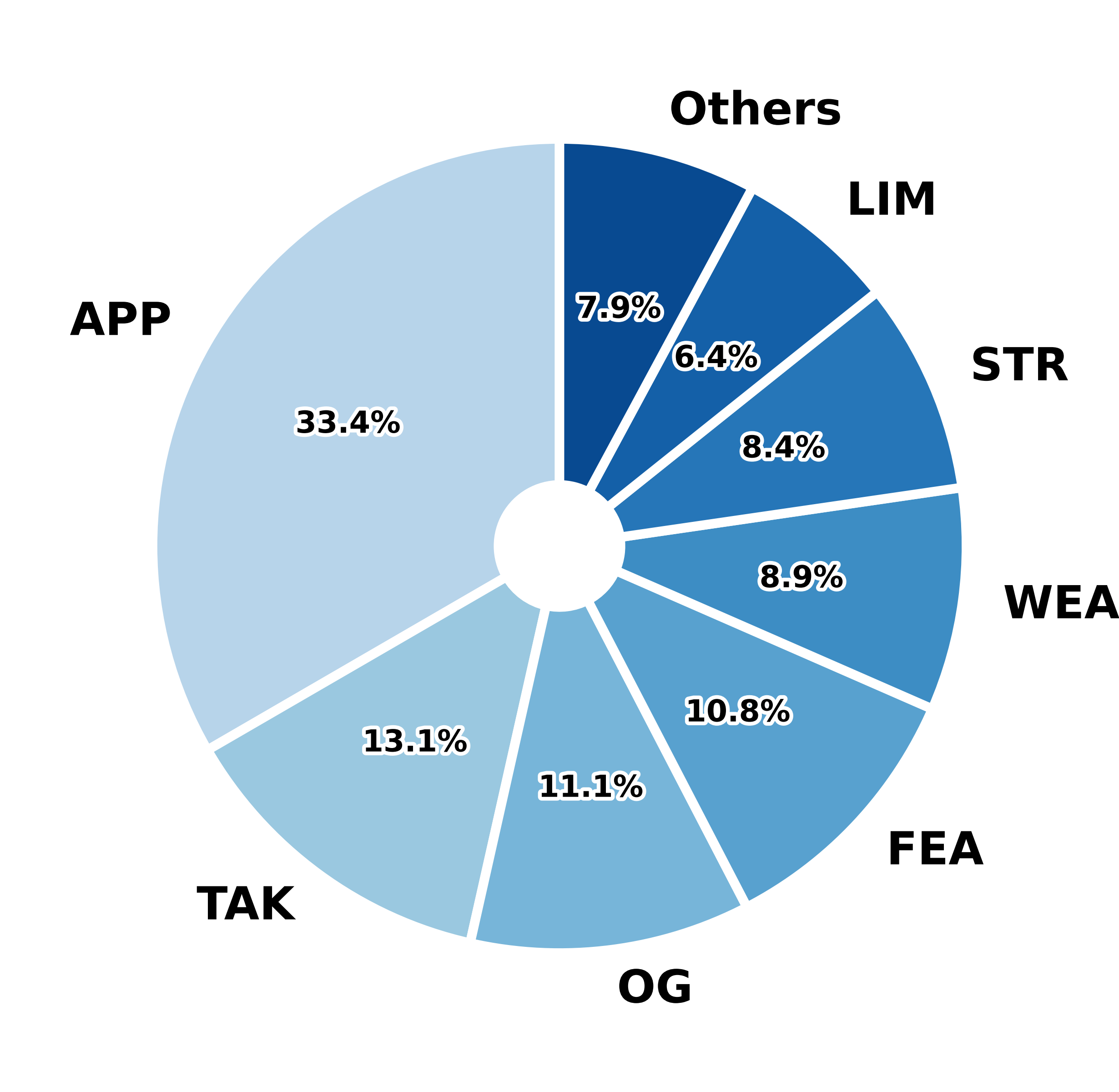}
        \caption{Overlap Nugget: Nugget Type}
        \label{fig:overlap_nugget_type}
    \end{subfigure}\hfill
    \begin{subfigure}{0.32\linewidth}
        \centering
        \includegraphics[width=\linewidth]{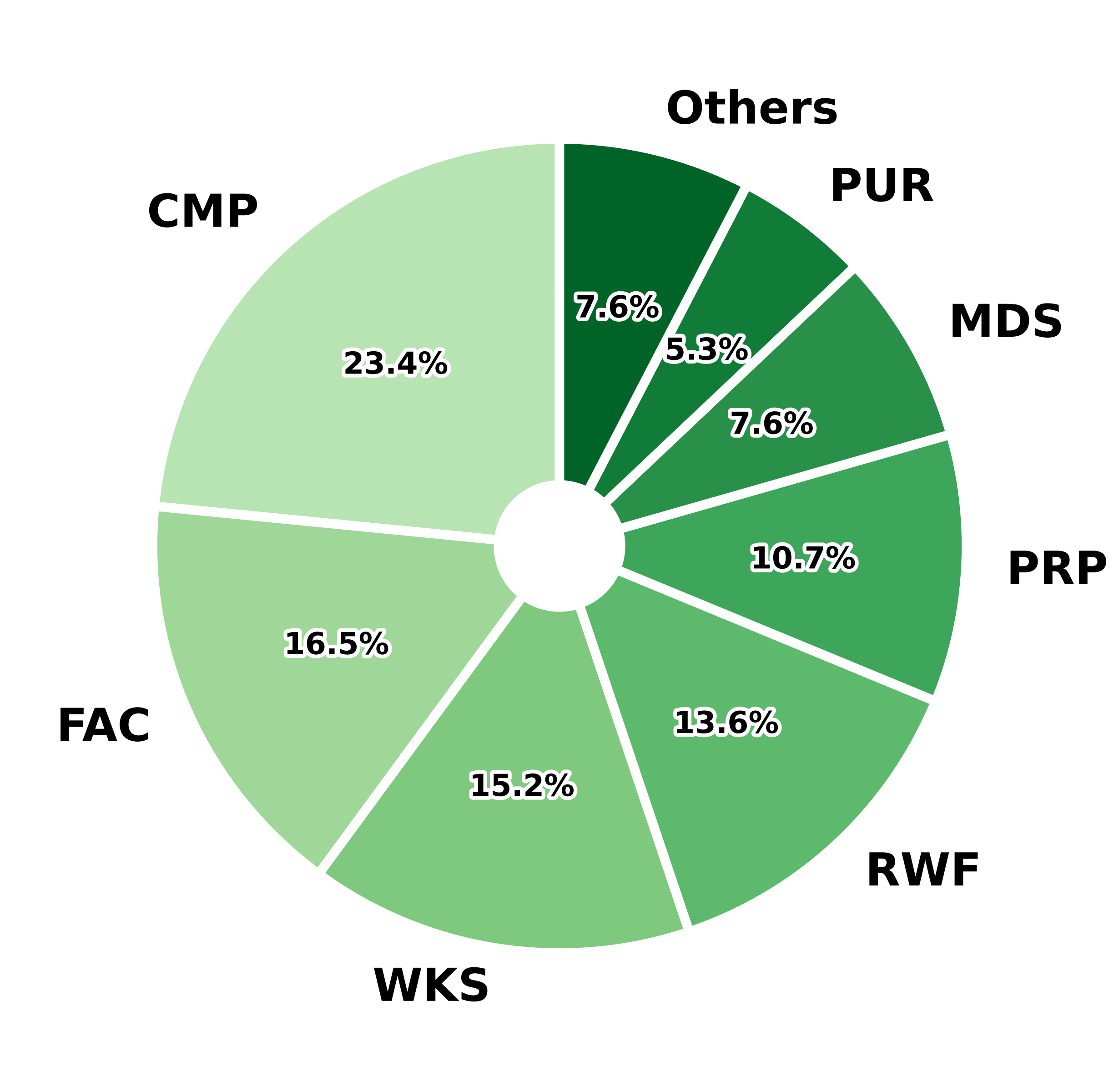}
        \caption{Overlap Nugget: Event Type}
        \label{fig:overlap_event_type}
    \end{subfigure}\hfill
    \begin{subfigure}{0.32\linewidth}
        \centering
        \includegraphics[width=\linewidth]{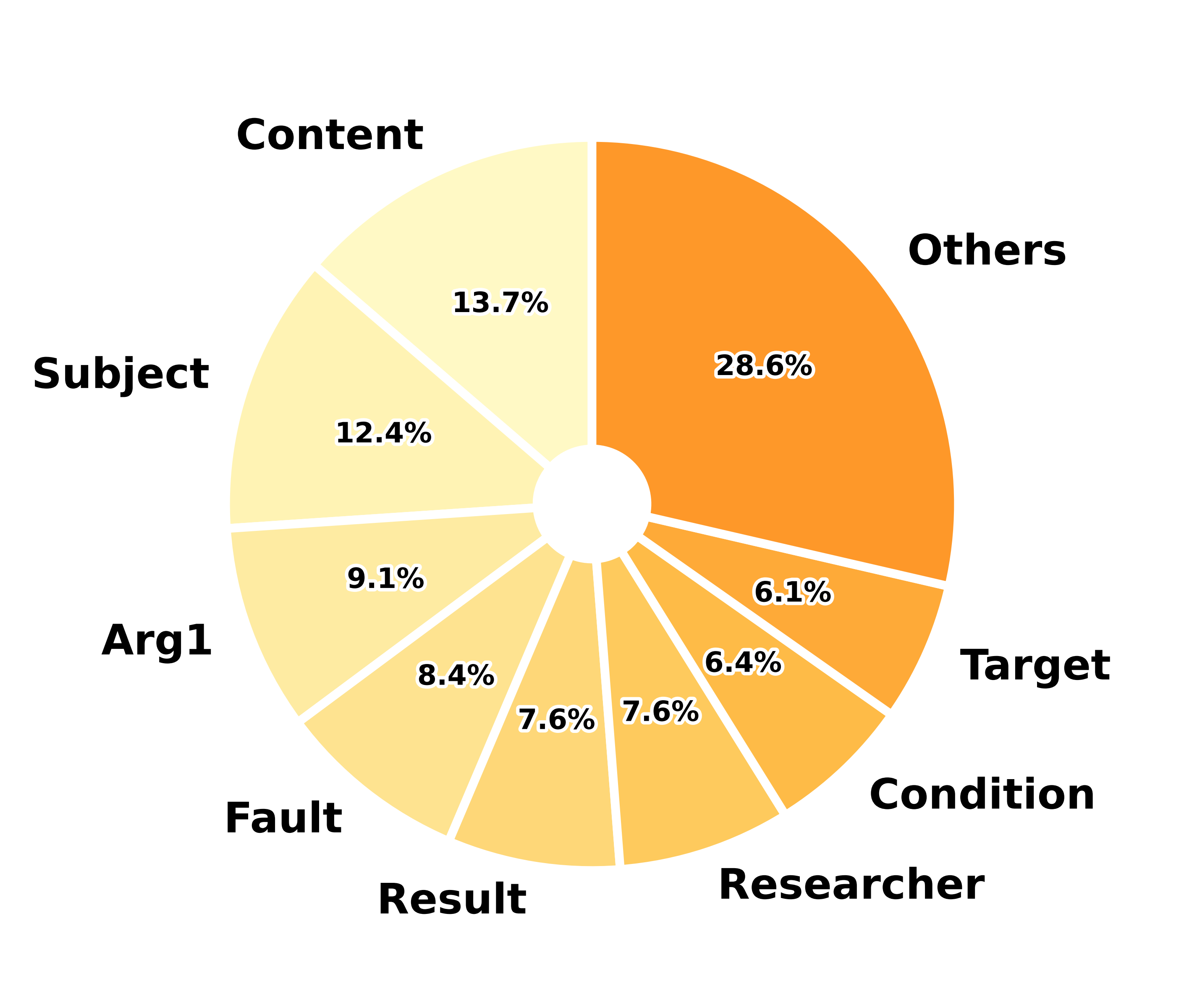}
        \caption{Overlap Nugget: Arg Type}
        \label{fig:overlap_argument_type}
    \end{subfigure}

    % ---------- Row 3 ----------
    \begin{subfigure}{0.32\linewidth}
        \centering
        \includegraphics[width=\linewidth]{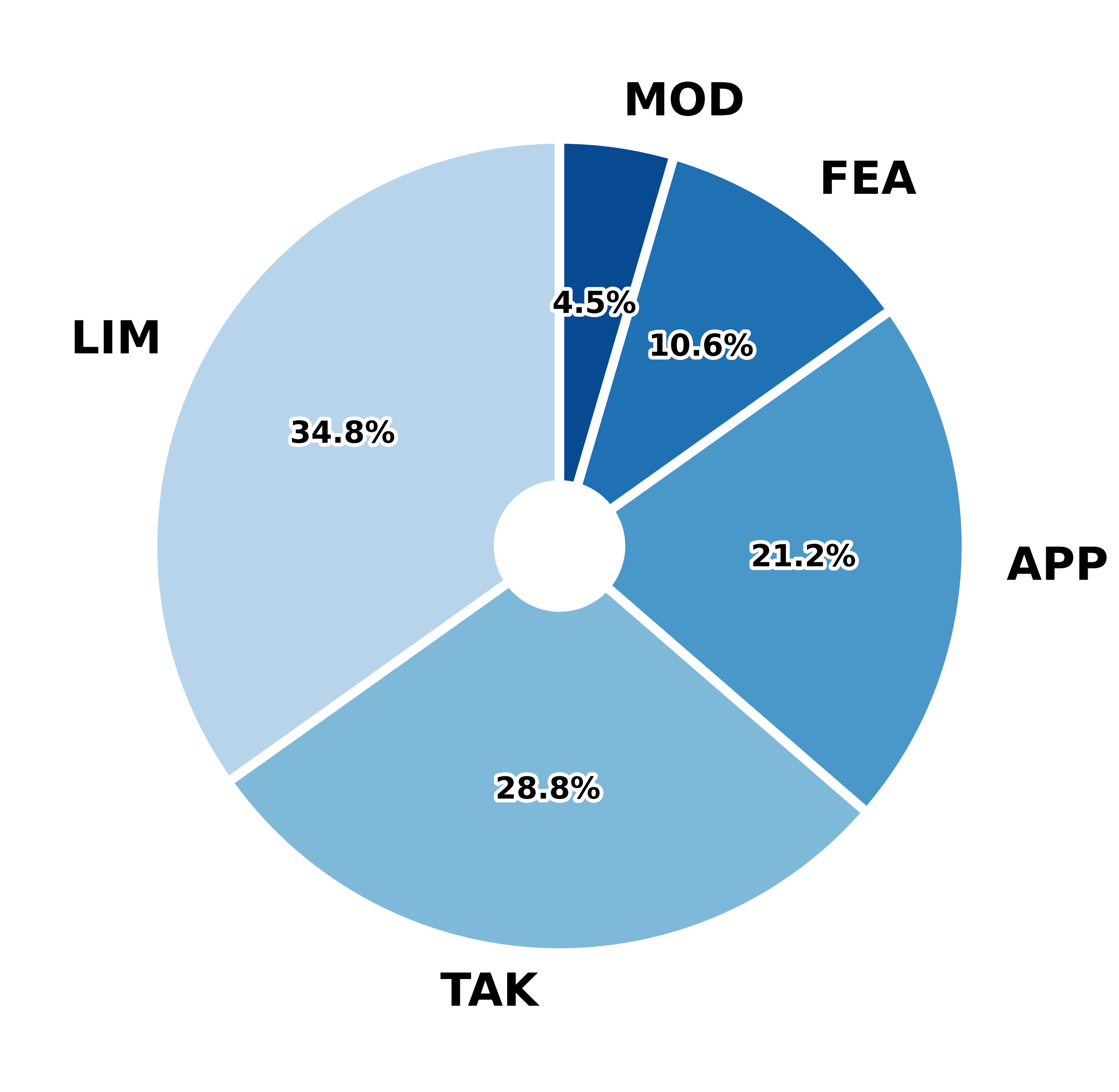}
        \caption{Reverse-Order Nugget: Nugget Type}
        \label{fig:reverseOrder_nugget_type}
    \end{subfigure}\hfill
    \begin{subfigure}{0.32\linewidth}
        \centering
        \includegraphics[width=\linewidth]{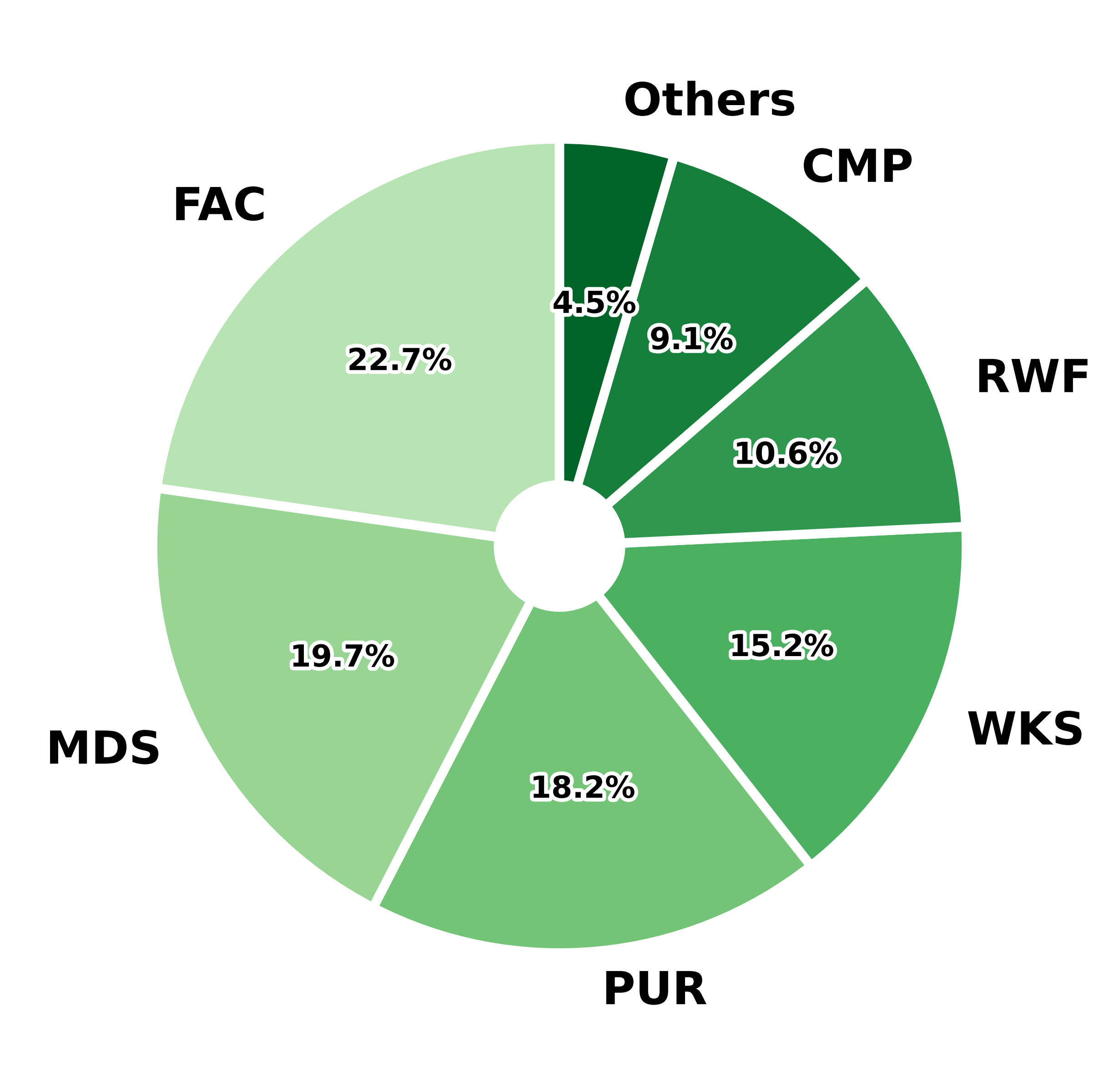}
        \caption{Reverse-Order Nugget: Event Type}
        \label{fig:reverseOrder_event_type}
    \end{subfigure}\hfill
    \begin{subfigure}{0.32\linewidth}
        \centering
        \includegraphics[width=\linewidth]{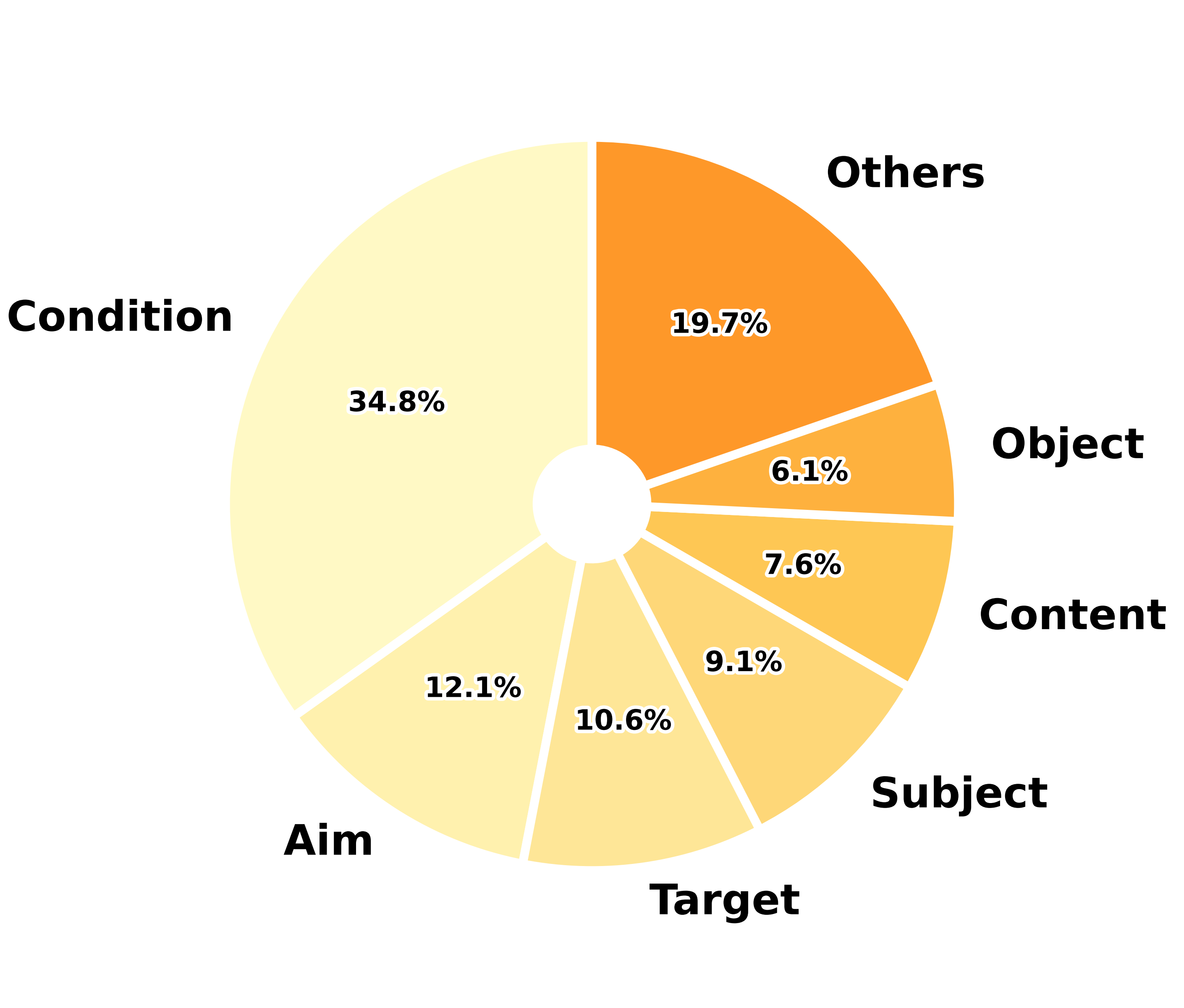}
        \caption{Reverse-Order Nugget: Arg Type}
        \label{fig:reverseOrder_argument_type}
    \end{subfigure}

    % ---------- Row 4 ----------
    \begin{subfigure}{0.32\linewidth}
        \centering
        \includegraphics[width=\linewidth]{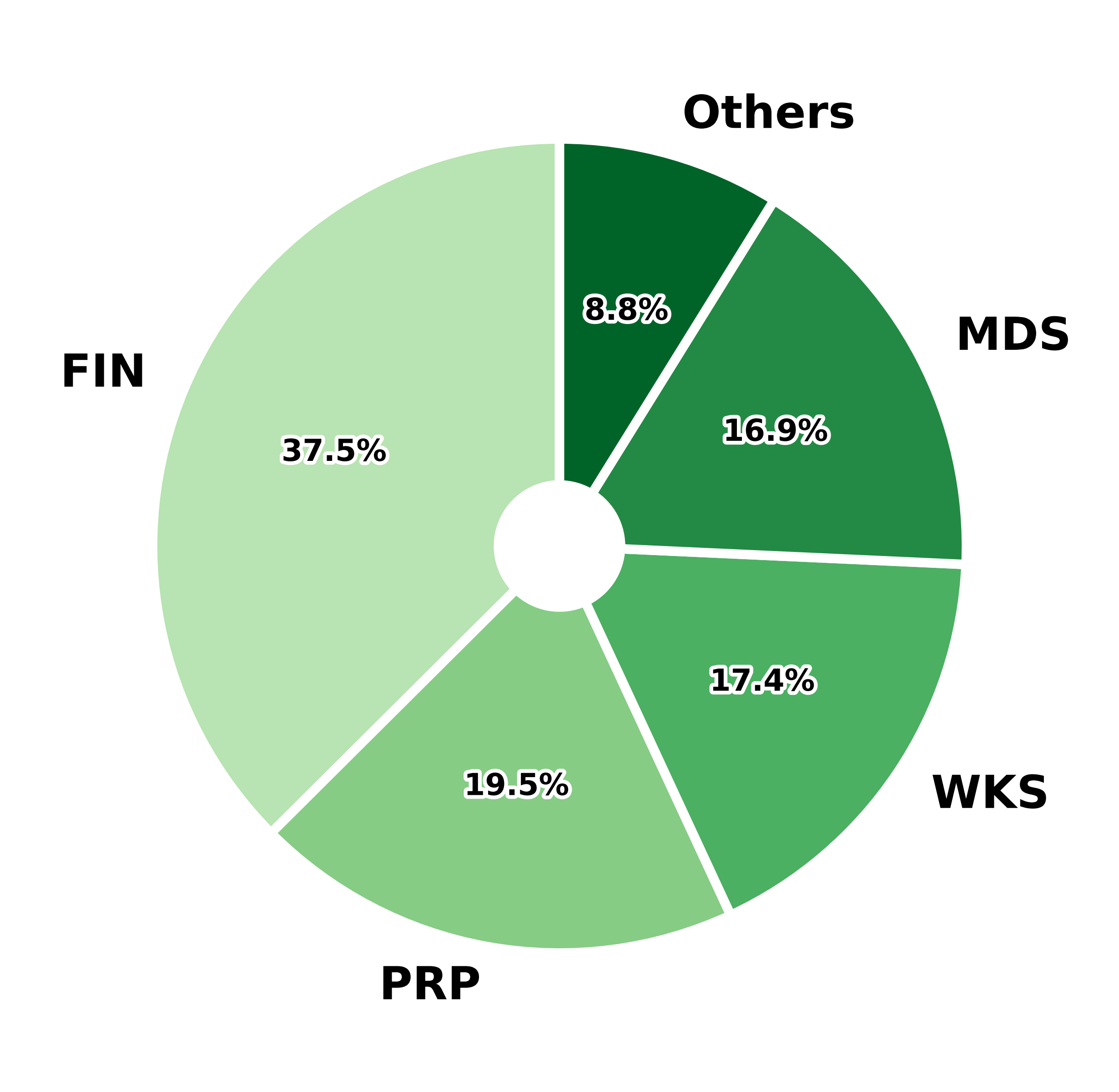}
        \caption{Sub-Event: Event Type}
        \label{fig:subEvent_event_type}
    \end{subfigure}\hfill
    \begin{subfigure}{0.32\linewidth}
        \centering
        \includegraphics[width=\linewidth]{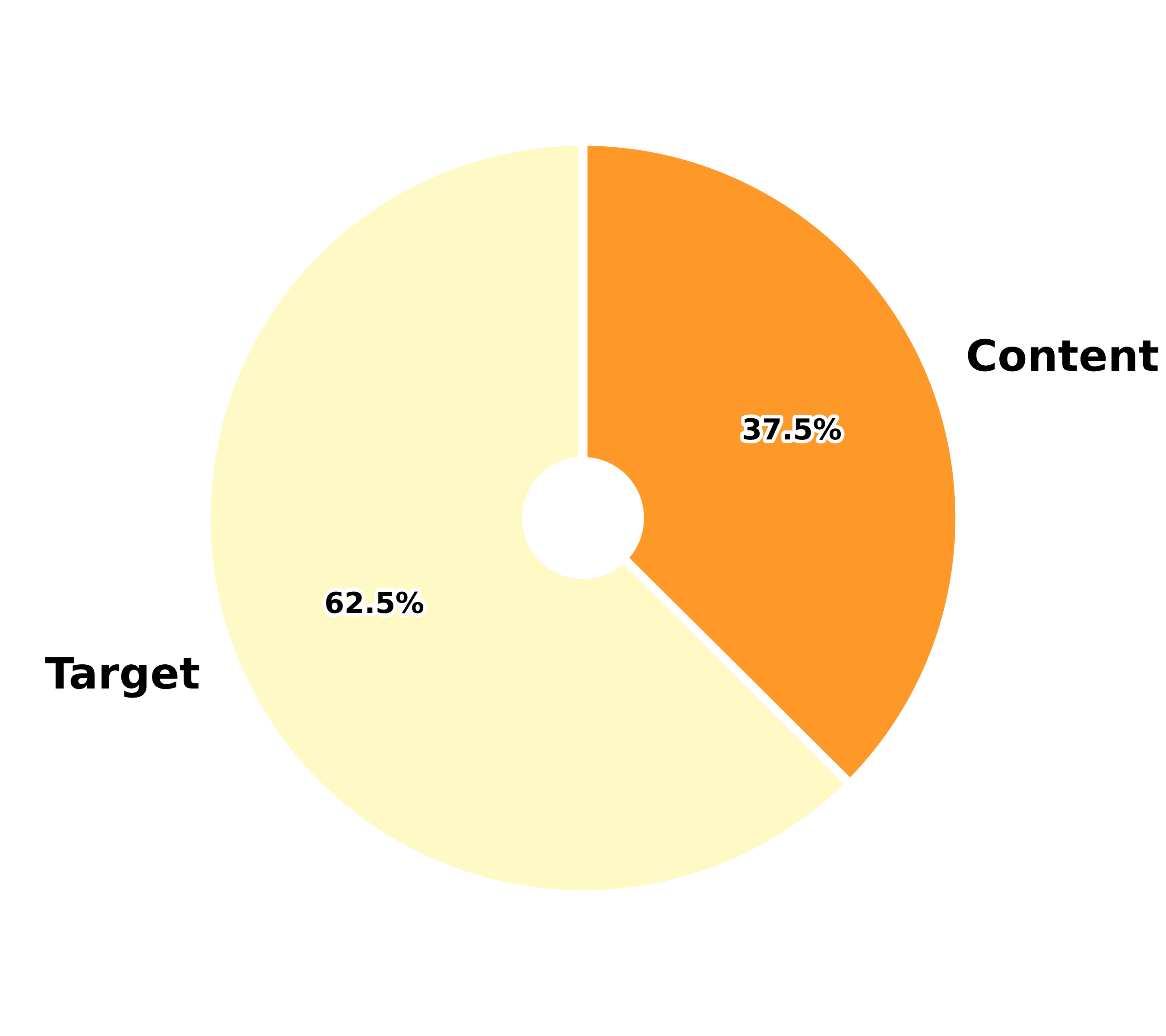}
        \caption{Sub-Event: Argument Type}
        \label{fig:subEvent_argument_type}
    \end{subfigure}\hfill
    \begin{subfigure}{0.32\linewidth}
        \centering
        \includegraphics[width=\linewidth]{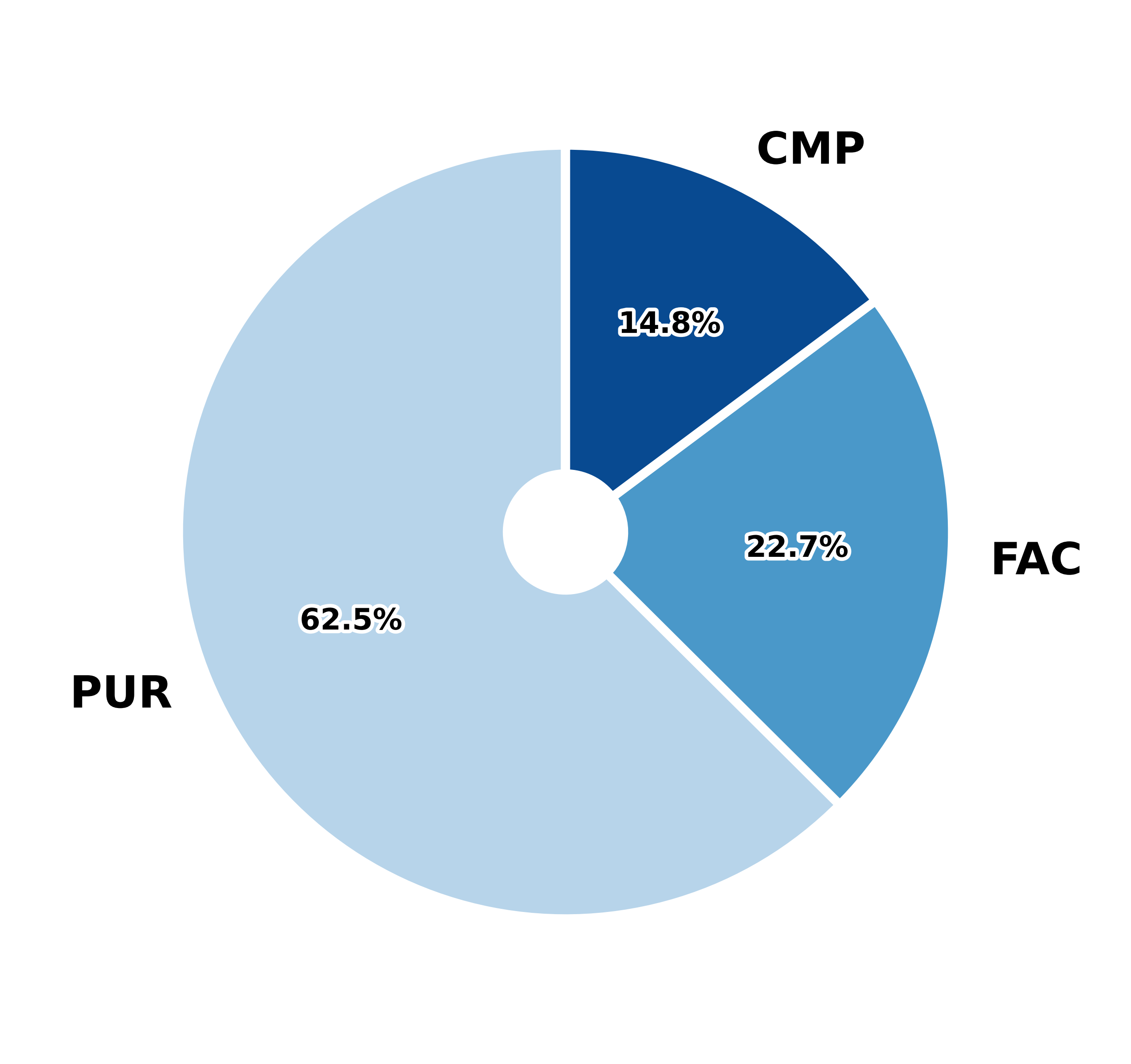}
        \caption{Sub-Event: Sub-Event Type}
        \label{fig:subEvent_subEvent_type}
    \end{subfigure}

    \caption{Distributions of Complex Nuggets and Events}
    \label{fig:12subfig}
\end{figure*}

\paragraph{Discontinuous Nugget Distribution.} Figures~\ref{fig:discontinuous_nugget_type}, \ref{fig:discontinuous_event_type}, and \ref{fig:discontinuous_argument_type} present the distribution of discontinuous nuggets over nugget types, event types and argument types, respectively.

\paragraph{Overlapping Nugget Distribution.}
Figures~\ref{fig:overlap_nugget_type}, \ref{fig:overlap_event_type}, and \ref{fig:overlap_argument_type} present the distribution of overlapping nuggets over nugget types, event types and argument types, respectively.

\paragraph{Reverse-order Nugget Distribution.}
Figures~\ref{fig:reverseOrder_nugget_type}, \ref{fig:reverseOrder_event_type}, and \ref{fig:reverseOrder_argument_type} present the distribution of reverse-order nuggets over nugget types, event types and argument types, respectively.

\paragraph{Sub-event Distribution.}
Figures~\ref{fig:subEvent_event_type}, \ref{fig:subEvent_argument_type}, and \ref{fig:subEvent_subEvent_type} present the distribution of sub-events over event types, argument types and sub-event types, respectively.

\section{Dataset Annotation Protocol and Reproducibility Details}
\label{sec:construction_details}
To facilitate reproducibility and provide practical guidance for future research, we detail the annotation protocol of SciEvents, which begins with the finalization of the schema and ends with the completion of the dataset.

With the defined schema of SciEvents, we hire a professional annotation company to support the annotation work of SciEvents. 
The entire annotation process is organized as a formal project by the annotation company, comprising the following four stages: project familiarization, annotator selection, annotator training, and formal annotation.
We provide the details of each stage as follows:
\paragraph{Project Familiarization Stage.} In this stage, we engage in in-depth discussions with senior staff from the annotation company. 
Specifically, we communicate closely with three senior staff members to clarify the input and output formats, data sources, schema, and annotation guidelines. 
These staff members are referred to as supervisors, as they will play leading roles in subsequent stages of the project.

To prepare the supervisors for leading the subsequent annotation process, we further conduct an iterative annotation and alignment procedure.
In each round, the three supervisors independently annotate the same set of 10 documents, followed by a joint discussion with our team to align their understanding of the annotation guidelines and refine the guidelines accordingly. 
This process is repeated until a consistent understanding is reached. 
In total, the procedure is conducted for approximately five rounds, with each supervisor annotating 50 documents.

\paragraph{Annotator Selection Stage.}
This stage is primarily led by the three supervisors. Specifically, they organize an internal project briefing within the company to recruit candidates for the pre-annotation phase, resulting in 21 participants. Based on two criteria, the basic understanding of the annotation guidelines and the ability to correctly identify event occurrences, they select 7 candidates as qualified annotators for the subsequent stages.

% \begin{figure*}[ht]
%   \centering
%   \includegraphics[width=\linewidth]{figs/annotation_screen.png}
%   \caption{A representative screenshot of the annotation tool developed by the annotation company (presented in Chinese). This screenshot illustrates the annotation interface, including an annotation panel on the left and a data management panel on the right. The core workflow includes selecting a document to annotate, adding a new event, selecting the event type, selecting spans in the document, assigning nugget types, assigning argument types, adding comments (optional), and submitting annotations.}
%   \label{fig:annotation_screen}
% \end{figure*}

% \begin{figure*}[ht]
%   \centering
%   \includegraphics[width=\linewidth]{figs/quality_inspection_screen.png}
%   \caption{A representative screenshot of the quality inspection tool developed by the annotation company (presented in Chinese). This screenshot illustrates the quality inspection interface, similar to the annotation interface. It allows inspectors to directly modify the annotations, provide feedback, and mark each document as approved or returned for revision.}
%   \label{fig:quality_inspection_screen}
% \end{figure*}

\paragraph{Annotator Training.} This stage is conducted in parallel with the formal annotation stage. It consists of three components: (1) \textit{One-on-one training:} Before formal annotation, each annotator receives approximately three days of one-on-one training from the supervisors, during which annotators are required to annotate at least 15 documents;
(2) \textit{On-demand support:} When annotators encounter ambiguous or difficult cases during annotation, they can directly consult the supervisors for clarification;
(3) \textit{Regular group sessions:} At least once per week, supervisors organize group sessions to summarize common issues identified during quality inspection and provide unified explanations.

\begin{figure*}[htbp]
  \centering

  \begin{subfigure}{\linewidth}
    \centering
    \includegraphics[width=\linewidth]{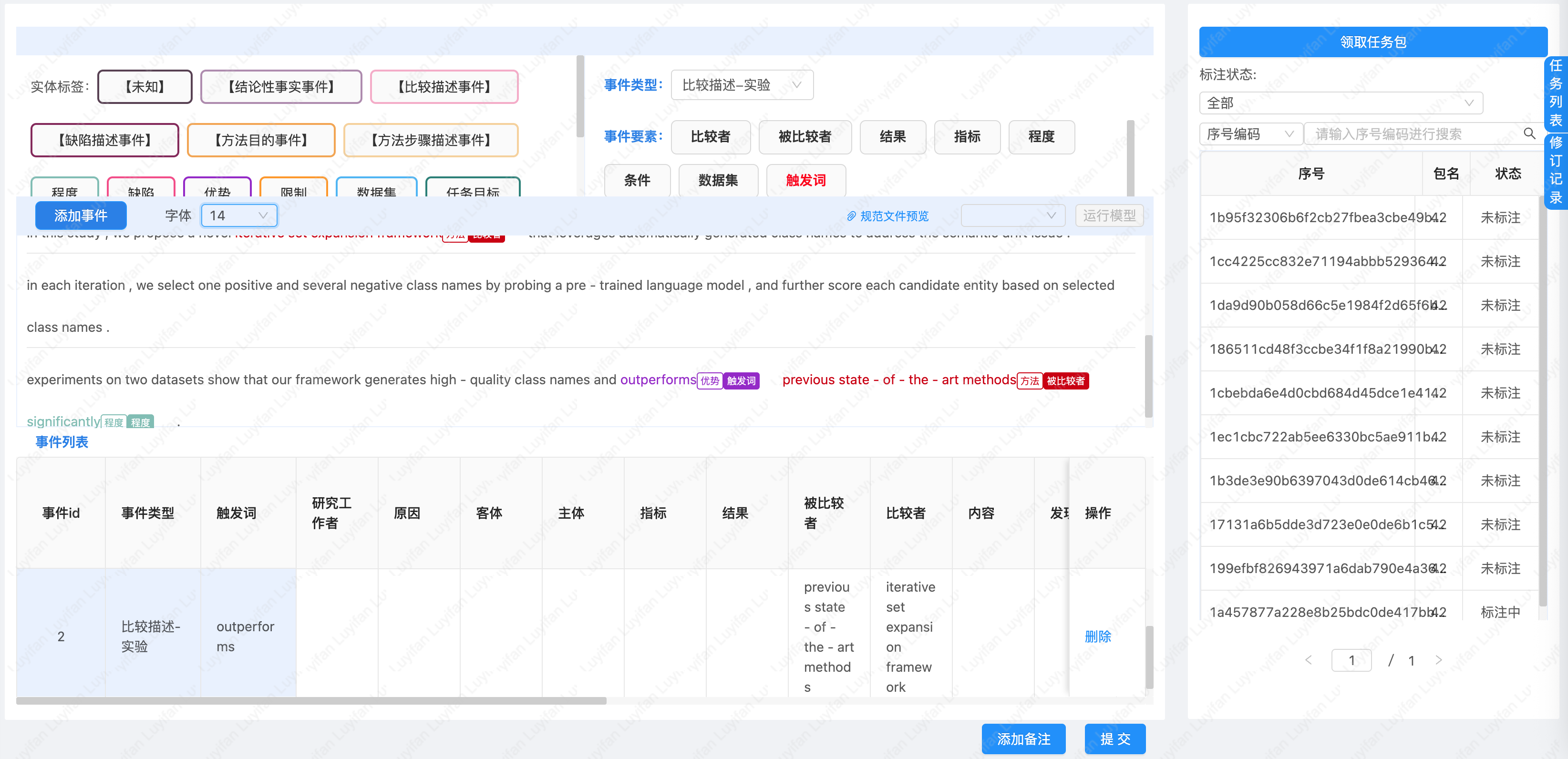}
    \caption{Annotation interface.}
    \label{fig:annotation_screen}
  \end{subfigure}

  \medskip

  \begin{subfigure}{\linewidth}
    \centering
    \includegraphics[width=\linewidth]{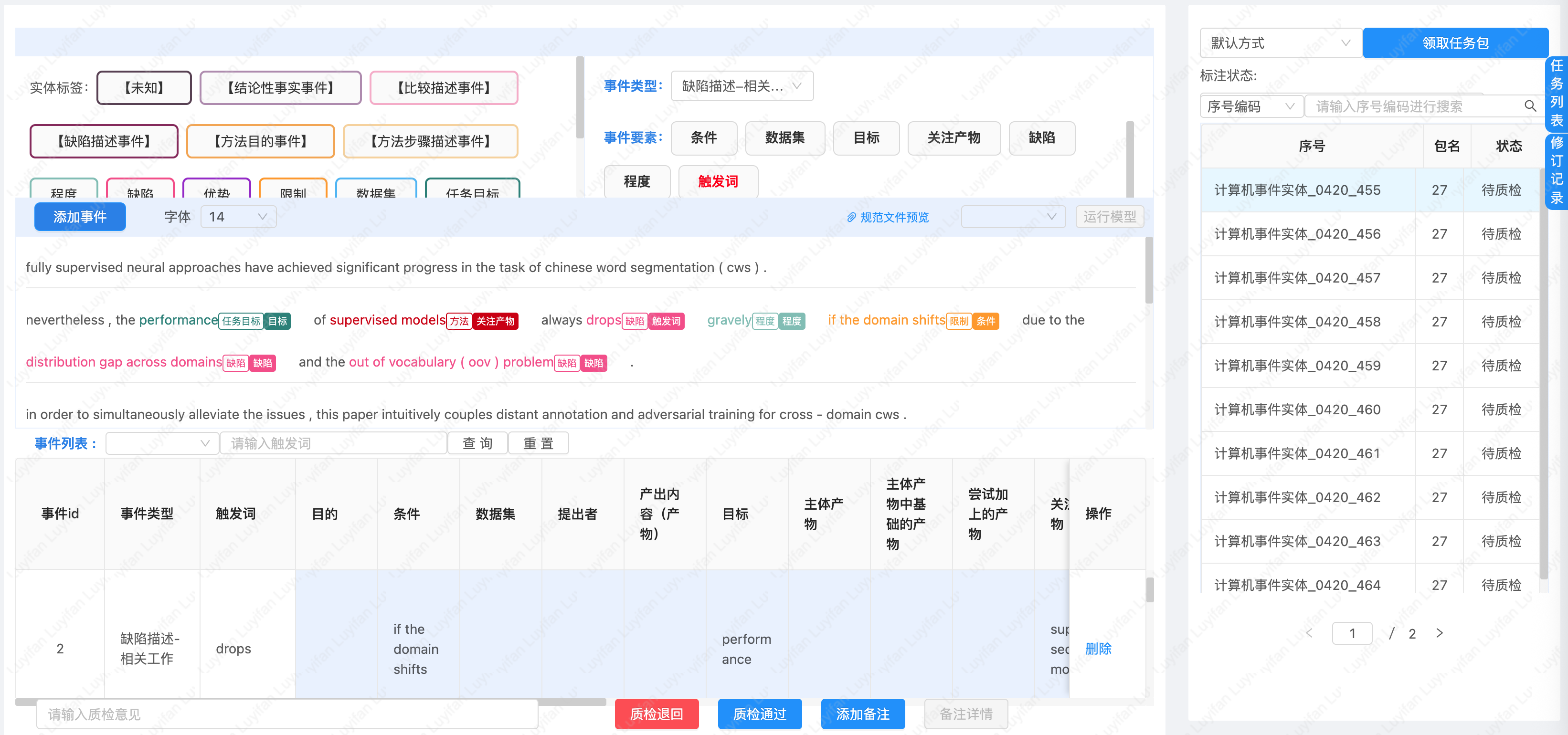}
    \caption{Quality inspection interface.}
    \label{fig:quality_inspection_screen}
  \end{subfigure}

  \caption{
  Representative screenshots of the annotation and quality inspection tools developed by the annotation company (presented in Chinese).
  (a) The annotation interface supports adding events, selecting event types, annotating spans, assigning nugget and argument types, and submitting annotations.
  (b) The quality inspection interface allows inspectors to modify annotations, provide feedback, and mark documents as approved or returned for revision.
  }

  \label{fig:annotation_and_qc}
\end{figure*}

\paragraph{Formal Annotation Stage.}
This stage includes both the annotation process and quality inspection. Most details are provided in the main paper. 
The annotation tool is independently developed and customized by the annotation company. 
Figure~\ref{fig:annotation_screen} and Figure~\ref{fig:quality_inspection_screen} illustrate representative interfaces for annotation and quality inspection, respectively.

\section{Dataset Annotation Remunerations}
During the official annotation stage, annotators spend approximately 364 minutes to annotate 19 consecutive documents, corresponding to an average of 19.1 minutes per document. Annotators are compensated at approximately \$4.5 per document, which is aligned with local wage standards and ensures fair remuneration for the annotation work.

\begin{table*}[htbp]
\centering
\scriptsize
\begin{tabular}{|c|c|c|c|c|c|}
\hline
\multicolumn{6}{|c|}{\textbf{Adaptive Compression of Word Embeddings}} \\
\hline
\multicolumn{6}{|c|}{\textbf{Document ID}: f63de5c23cce0cc5bb67d42ab12e7bed} \\
\hline
\multicolumn{6}{|p{0.95\linewidth}|}{\textbf{Abstract}: Distributed representations of words have been an indispensable \textbf{component}$_{\rm E1}$ for natural language processing (NLP) tasks. However, the \textbf{large memory footprint}$_{\rm E2}$ of word embeddings makes it challenging to deploy NLP models to memory-constrained devices (e.g., self-driving cars, mobile devices). In this paper, we \textbf{propose}$_{\rm E3}$ a novel method to \textbf{adaptively compress}$_{\rm E4}$ word embeddings. We fundamentally \textbf{follow}$_{\rm E5}$ a code-book approach that \textbf{represents}$_{\rm E6}$ words as discrete codes such as (8, 5, 2, 4). However, unlike prior works that assign the same length of codes to all words, we adaptively \textbf{assign}$_{\rm E8}$ different lengths of codes to each word by \textbf{learning}$_{\rm E7}$ downstream tasks. The proposed method works in two steps. First, each word directly learns to \textbf{select}$_{\rm E10}$ its code length in an end-to-end manner by \textbf{applying}$_{\rm E9}$ the Gumbel-softmax tricks. After selecting the code length, each word \textbf{learns}$_{\rm E12}$ discrete codes \textbf{through}$_{\rm E11}$ a neural network with a binary constraint. To \textbf{showcase}$_{\rm E14}$ the general applicability of the proposed method, we \textbf{evaluate}$_{\rm E13}$ the performance on four different downstream tasks. Comprehensive evaluation results clearly \textbf{show}$_{\rm E15}$ that our method is \textbf{effective}$_{\rm E16}$ and \textbf{makes}$_{\rm E17}$ the highly compressed word embeddings without hurting the task accuracy. Moreover, we \textbf{show}$_{\rm E18}$ that our model \textbf{assigns}$_{\rm E20}$ word to each code-book by \textbf{considering}$_{\rm E19}$ the significance of tasks.} \\
\hline
Event ID & Event Type & Trigger & Arg Type & Arg Text & Nugget Type \\
\hline
E1 & ITT & component & Target & natural language processing & TAK \\
\hline
E2 & RWF & large memory footprint & Concern & word embeddings & MOD \\
\hline
E3 & PRP & propose & Proposer & we & OG \\
 & & & Content & method & APP \\
 & & & Target & adaptively compress & E-PUR$^\dagger$ \\
\hline
E4 & PUR$^*$ & adaptively compress & Aim & word embeddings & MOD \\
\hline
E5 & WKS & follow & Researcher & we & OG \\
 & & & Content & code - book approach & APP \\
 & & & Target & represents & E-PUR$^\dagger$ \\
\hline
E6 & PUR$^*$ & represents & Aim & words & FEA \\
 & & & Condition & as discrete codes & LIM \\
\hline
E7 & WKS & learning & Content & downstream tasks & TAK \\
 & & & Researcher & we & OG \\
 & & & Target & assign & E-PUR$^\dagger$ \\
\hline
E8 & PUR$^*$ & assign & Aim & different lengths of codes & FEA \\
 & & & Condition & to each word & LIM \\
\hline
E9 & MDS & applying & Target & select & E-PUR$^\dagger$ \\
 & & & TriedComponent & gumbel - softmax tricks & APP \\
 & & & BaseComponent & word & FEA \\
\hline
E10 & PUR$^*$ & select & Aim & code length & TAK \\
 & & & Condition & in an end - to - end manner & LIM \\
\hline
E11 & MDS & through & Target & learns & E-PUR$^\dagger$ \\
 & & & BaseComponent & word & FEA \\
 & & & TriedComponent & neural network with a binary constraint & APP \\
 & & & Condition & after selecting the code length & LIM \\
\hline
E12 & PUR$^*$ & learns & Aim & discrete codes & TAK \\
\hline
E13 & WKS & evaluate & Researcher & we & OG \\
 & & & Content & performance & TAK \\
 & & & Condition & on four different downstream tasks & LIM \\
 & & & Target & showcase & E-PUR$^\dagger$ \\
\hline
E14 & PUR$^*$ & showcase & Aim & general applicability & TAK \\
\hline
E15 & FIN & show & Content & effective & E-FAC$^\dagger$ \\
 & & & Content & makes & E-FAC$^\dagger$ \\
\hline
E16 & FAC$^*$ & effective & Subject & method & APP \\
\hline
E17 & FAC$^*$ & makes & Condition & without hurting the task accuracy & LIM \\
 & & & Subject & method & APP \\
 & & & Object & highly compressed word embeddings & STR \\
\hline
E18 & FIN & show & Finder & we & OG \\
 & & & Content & considering & E-FAC$^\dagger$ \\
\hline
E19 & FAC$^*$ & considering & Object & significance of tasks & TAK \\
 & & & Target & assigns & E-PUR$^\dagger$ \\
 & & & Subject & model & APP \\
\hline
E20 & PUR & assigns & Aim & word & FEA \\
 & & & Condition & to each code - book & LIM \\
\hline
\end{tabular}
\caption{Event extraction annotations for the paper \textit{Adaptive Compression of Word Embeddings}. $^*$ indicates that the event is a sub-event; $^\dagger$ indicates that the argument is a sub-event argument (nugget\_type starts with E-).}
\label{tab:case_f63de5c2}
\end{table*}

\begin{table*}[ht]
    \small
    \centering
    \begin{threeparttable}
        \begin{tabular}{c|l|cccccc}
            \toprule  
            \multicolumn{2}{c}{\multirow{2}{*}{\bf Model}} & \multicolumn{3}{c}{TI(\%)} & \multicolumn{3}{c}{TC(\%)} \\
            \multicolumn{2}{c}{~} & P & R & F1 & P & R & F1 \\
            \midrule
            \multirow{1}{*}{Global} &
                OneIE & 74.18$_{\pm0.24}$ & 77.33$_{\pm0.38}$ & 75.72$_{\pm0.14}$ & 61.68$_{\pm0.06}$ & 64.23$_{\pm0.42}$ & 62.93$_{\pm0.17}$ \\

            \midrule
            \multirow{2}{*}{Discriminative} 
            ~ & EEQA & 77.04$_{\pm2.13}$ & 72.96$_{\pm3.19}$ & 74.85$_{\pm0.78}$ & 64.03$_{\pm1.87}$ & 60.53$_{\pm2.64}$ & 62.15$_{\pm0.73}$ \\
            ~ & Tagprime & 75.72$_{\pm1.16}$ & 71.04$_{\pm1.98}$ & 73.27$_{\pm0.52}$ & 64.85$_{\pm1.54}$ & 61.36$_{\pm1.05}$ & 63.03$_{\pm0.20}$ \\

            \midrule
            \multirow{2}{*}{Generative} &
                DEGREE & 64.98$_{\pm1.72}$ & 66.53$_{\pm1.08}$ & 65.72$_{\pm0.70}$ & 51.93$_{\pm1.80}$ & 55.27$_{\pm0.42}$ & 53.52$_{\pm0.75}$ \\
            ~ & KnowCoder & 69.83$_{\pm0.58}$ & 69.95$_{\pm0.86}$ & 69.88$_{\pm0.61}$ & 52.02$_{\pm0.30}$ & 52.03$_{\pm0.55}$ & 52.02$_{\pm0.34}$ \\
            
            \midrule
            \multicolumn{2}{l|}{EXCEEDS (Ours)} & 73.90$_{\pm0.78}$ & 76.76$_{\pm1.44}$ & 75.29$_{\pm0.32}$ & 62.28$_{\pm0.78}$ & 65.31$_{\pm1.14}$ & 63.74$_{\pm0.14}$\\
            \multicolumn{2}{l|}{~~-- Contextual} & 73.02$_{\pm1.38}$ & 77.75$_{\pm1.38}$ & 75.29$_{\pm0.21}$ & 61.34$_{\pm1.46}$ & 65.74$_{\pm1.26}$ & 63.44$_{\pm0.67}$ \\
            \multicolumn{2}{l|}{~~-- Grid Refiner} & 73.73$_{\pm0.73}$ & 77.09$_{\pm1.01}$ & 75.36$_{\pm0.27}$ & 61.76$_{\pm0.97}$ & 65.16$_{\pm1.02}$ & 63.41$_{\pm0.67}$ \\
            % ~ & old-ex & 72.31 & \bf 79.92 & \bf 75.92 & 60.12 & \bf 66.41 & \bf 63.11 \\ 
            \bottomrule
        \end{tabular}
    \caption{Precision, recall and F1-score (\%) of trigger identification (TI) and trigger classification (TC) on SciEvents. EAE-only models are not presented.}
    \label{tab:prf_trigger}
    \end{threeparttable}
\end{table*}

\begin{table*}[ht]
    \small
    \centering
    \begin{threeparttable}
    \begin{tabular}{c|l|cccccc}
        \toprule  
        \multicolumn{2}{c}{\multirow{2}{*}{\bf Model}} & \multicolumn{3}{c}{AI(\%)} & \multicolumn{3}{c}{AC(\%)} \\
        \multicolumn{2}{c}{~} & P & R & F1 & P & R & F1 \\
        \midrule
        \multirow{2}{*}{Global} &
            OneIE & 28.71$_{\pm1.80}$ & 32.09$_{\pm1.90}$ & 30.30$_{\pm1.85}$ & 27.29$_{\pm1.73}$ & 30.49$_{\pm1.82}$ & 28.81$_{\pm1.77}$ \\
        ~ & ScentedEAE & 50.73$_{\pm1.66}$ & 29.07$_{\pm4.12}$ & 36.70$_{\pm2.99}$ & 49.48$_{\pm2.32}$ & 28.29$_{\pm3.61}$ & 35.74$_{\pm2.41}$ \\

        \midrule
        \multirow{4}{*}{Discriminative} & 
         EEQA & 32.65$_{\pm0.97}$ & 44.93$_{\pm2.45}$ & 37.75$_{\pm0.46}$ & 30.75$_{\pm0.78}$ & 42.55$_{\pm2.53}$ & 35.64$_{\pm0.59}$ \\
        ~ & PAIE & 46.86$_{\pm0.26}$ & 41.34$_{\pm0.59}$ & 43.92$_{\pm0.22}$ & 44.82$_{\pm0.31}$ & 39.63$_{\pm0.65}$ & 42.06$_{\pm0.34}$ \\
        ~ & Tagprime & 47.79$_{\pm0.71}$ & 41.95$_{\pm0.47}$ & 44.67$_{\pm0.13}$ & 45.67$_{\pm0.55}$ & 40.09$_{\pm0.67}$ & 42.69$_{\pm0.32}$ \\
        ~ & DEEIA & 43.06$_{\pm2.41}$ & 29.33$_{\pm0.37}$ & 34.86$_{\pm0.81}$ & 41.05$_{\pm2.23}$ & 28.06$_{\pm0.36}$ & 33.30$_{\pm0.75}$ \\

        \midrule
        \multirow{3}{*}{Generative} &
            BartGen & 41.21$_{\pm1.34}$ & 38.60$_{\pm0.58}$ & 39.85$_{\pm0.52}$ & 39.10$_{\pm1.21}$ & 36.63$_{\pm0.48}$ & 37.81$_{\pm0.39}$ \\
        ~ & DEGREE & 29.03$_{\pm1.33}$ & 27.81$_{\pm0.50}$ & 28.40$_{\pm0.88}$ & 26.87$_{\pm1.21}$ & 25.79$_{\pm0.43}$ & 26.32$_{\pm0.79}$ \\
        ~ & KnowCoder & 37.82$_{\pm0.17}$ & 33.00$_{\pm0.17}$ & 35.24$_{\pm0.11}$ & 35.89$_{\pm0.35}$ & 31.29$_{\pm0.24}$ & 33.43$_{\pm0.27}$ \\
            
        \midrule 
        \multicolumn{2}{l|}{EXCEEDS (Ours)} & 50.86$_{\pm2.68}$ & 40.48$_{\pm1.92}$ & 44.97$_{\pm0.28}$ & 48.82$_{\pm2.81}$ & 38.91$_{\pm1.66}$ & 43.20$_{\pm0.29}$\\
        \multicolumn{2}{l|}{~~-- Contextual} & 48.64$_{\pm2.16}$ & 40.40$_{\pm1.70}$ & 44.07$_{\pm0.51}$ & 46.46$_{\pm2.35}$ & 38.68$_{\pm1.43}$ & 42.14$_{\pm0.39}$ \\
        \multicolumn{2}{l|}{~~-- Grid Refiner} & 49.64$_{\pm1.24}$ & 40.03$_{\pm0.94}$ & 44.30$_{\pm0.51}$ & 47.51$_{\pm1.37}$ & 38.39$_{\pm0.87}$ & 42.44$_{\pm0.59}$ \\
        
        % ~ & old-ex & \bf 46.94 & 41.82 & \bf 44.23 & \bf 44.43 & 39.55 & \bf 41.85 \\ 
        \bottomrule
    \end{tabular}
    \caption{Precision, recall and F1-score (\%) of argument identification (AI) and argument classification (AC) on SciEvents. For EAE-only models, trigger predictions are derived from Tagprime.}
    \label{tab:prf_argument}
    \end{threeparttable}
\end{table*}

\begin{table*}[ht]
    \small
    \centering
    \begin{threeparttable}
    \begin{tabular}{c|l|ccc}
        \toprule  
        \multicolumn{2}{c}{\multirow{2}{*}{\bf Model}} & \multicolumn{3}{c}{EC(\%)} \\
        \multicolumn{2}{c}{~} & P & R & F1 \\
        \midrule
        \multirow{2}{*}{Global} &
            OneIE & 35.26$_{\pm1.48}$ & 39.85$_{\pm1.61}$ & 37.41$_{\pm1.51}$ \\
        ~ & ScentedEAE & 51.21$_{\pm2.24}$ & 30.12$_{\pm2.32}$ & 37.88$_{\pm2.10}$ \\

        \midrule
        \multirow{4}{*}{Discriminative} & 
        EEQA & 44.45$_{\pm3.77}$ & 45.57$_{\pm2.38}$ & 44.81$_{\pm1.35}$ \\
        ~ & PAIE & 49.98$_{\pm2.11}$ & 44.70$_{\pm0.06}$ & 47.17$_{\pm0.92}$ \\
        ~ & Tagprime & 50.22$_{\pm1.42}$ & 45.53$_{\pm1.46}$ & 47.72$_{\pm0.32}$ \\
        ~ & DEEIA & 40.64$_{\pm3.35}$ & 27.53$_{\pm0.79}$ & 32.80$_{\pm1.67}$ \\

        \midrule
        \multirow{3}{*}{Generative} &
            BartGen & 46.93$_{\pm1.37}$ & 39.30$_{\pm0.87}$ & 42.75$_{\pm0.11}$ \\
        ~ & DEGREE & 39.60$_{\pm1.01}$ & 23.55$_{\pm0.86}$ & 29.53$_{\pm0.96}$ \\
        ~ & KnowCoder & 41.03$_{\pm0.94}$ & 29.85$_{\pm1.22}$ & 34.54$_{\pm0.82}$ \\
            
        \midrule 
        \multicolumn{2}{l|}{EXCEEDS (Ours)} & 48.28$_{\pm1.05}$ & 48.29$_{\pm1.22}$ & 48.25$_{\pm0.10}$\\
        \multicolumn{2}{l|}{~~-- Contextual} & 47.05$_{\pm1.10}$ & 48.32$_{\pm2.13}$ & 47.64$_{\pm0.85}$ \\
        \multicolumn{2}{l|}{~~-- Grid Refiner} & 47.33$_{\pm0.86}$ & 48.80$_{\pm1.80}$ & 48.04$_{\pm1.07}$ \\
        % ~ & old-ex & 44.45 & \bf 51.25 & \bf 47.60 \\ 
        \bottomrule
    \end{tabular}
    \caption{Precision, recall and F1-score (\%) of event correlation (EC) on SciEvents. For EAE-only models, trigger predictions are derived from Tagprime.}
    \label{tab:prf_event}
    \end{threeparttable}
\end{table*}

\begin{table*}[ht]
\centering
\small
\begin{tabular}{lcccc}
\toprule
\textbf{Nugget Type} & \textbf{Train} & \textbf{Develop} & \textbf{Test} & \textbf{Total} \\
\midrule
APP & 8,860 (22.16\%) & 1,068 (21.38\%) & 1,146 (22.50\%) & 11,074 (22.12\%) \\
TAK & 8,219 (20.56\%) & 993 (19.88\%) & 1,042 (20.46\%) & 10,254 (20.48\%) \\
FEA & 6,147 (15.38\%) & 801 (16.03\%) & 764 (15.00\%) & 7,712 (15.40\%) \\
OG & 4,663 (11.67\%) & 567 (11.35\%) & 564 (11.07\%) & 5,794 (11.57\%) \\
LIM & 3,933 (9.84\%) & 536 (10.73\%) & 505 (9.92\%) & 4,974 (9.94\%) \\
STR & 1,986 (4.97\%) & 247 (4.94\%) & 249 (4.89\%) & 2,482 (4.96\%) \\
WEA & 1,779 (4.45\%) & 216 (4.32\%) & 246 (4.83\%) & 2,241 (4.48\%) \\
DST & 1,736 (4.34\%) & 219 (4.38\%) & 238 (4.67\%) & 2,193 (4.38\%) \\
MOD & 1,652 (4.13\%) & 240 (4.80\%) & 206 (4.04\%) & 2,098 (4.19\%) \\
DEG & 998 (2.50\%) & 109 (2.18\%) & 133 (2.61\%) & 1,240 (2.48\%) \\
\midrule
\textbf{Total} & \textbf{39,973} & \textbf{4,996} & \textbf{5,093} & \textbf{50,062} \\
\bottomrule
\end{tabular}
\caption{Distribution of Nugget Types across Train, Develop and Test Splits}
\label{tab:nugget_distribution}
\end{table*}

\begin{table*}[ht]
\centering
\small
\begin{tabular}{lcccc}
\toprule
\textbf{Event Type} & \textbf{Train} & \textbf{Develop} & \textbf{Test} & \textbf{Total} \\
\midrule
PUR & 3,320 (17.04\%) & 395 (16.16\%) & 416 (16.99\%) & 4,131 (16.94\%) \\
WKS & 2,733 (14.02\%) & 354 (14.48\%) & 332 (13.56\%) & 3,419 (14.02\%) \\
FAC & 2,528 (12.97\%) & 313 (12.80\%) & 324 (13.24\%) & 3,165 (12.98\%) \\
MDS & 2,385 (12.24\%) & 303 (12.39\%) & 300 (12.25\%) & 2,988 (12.26\%) \\
PRP & 2,094 (10.75\%) & 235 (9.61\%) & 257 (10.50\%) & 2,586 (10.61\%) \\
RWF & 1,549 (7.95\%) & 201 (8.22\%) & 205 (8.37\%) & 1,955 (8.02\%) \\
CMP & 1,536 (7.88\%) & 192 (7.85\%) & 195 (7.97\%) & 1,923 (7.89\%) \\
FIN & 1,480 (7.59\%) & 192 (7.85\%) & 175 (7.15\%) & 1,847 (7.58\%) \\
ITT & 1,400 (7.18\%) & 190 (7.77\%) & 174 (7.11\%) & 1,764 (7.24\%) \\
RWS & 463 (2.38\%) & 70 (2.86\%) & 70 (2.86\%) & 603 (2.47\%) \\
\midrule
\textbf{Total} & \textbf{19,488} & \textbf{2,445} & \textbf{2,448} & \textbf{24,381} \\
\bottomrule
\end{tabular}
\caption{Distribution of Event Types across Train, Develop and Test Splits}
\label{tab:event_distribution}
\end{table*}

\begin{table*}[ht]
\centering
\small
\begin{tabular}{lcccc}
\toprule
\textbf{Argument Type} & \textbf{Train} & \textbf{Develop} & \textbf{Test} & \textbf{Total} \\
\midrule
Content & 7,010 (15.55\%) & 877 (15.61\%) & 853 (14.92\%) & 8,740 (15.49\%) \\
Target & 6,982 (15.49\%) & 844 (15.02\%) & 876 (15.32\%) & 8,702 (15.43\%) \\
Condition & 3,919 (8.69\%) & 534 (9.51\%) & 506 (8.85\%) & 4,959 (8.79\%) \\
Aim & 3,574 (7.93\%) & 424 (7.55\%) & 443 (7.75\%) & 4,441 (7.87\%) \\
BaseComponent & 2,858 (6.34\%) & 377 (6.71\%) & 366 (6.40\%) & 3,601 (6.38\%) \\
Subject & 2,662 (5.91\%) & 326 (5.80\%) & 365 (6.38\%) & 3,353 (5.94\%) \\
TriedComponent & 2,143 (4.75\%) & 274 (4.88\%) & 279 (4.88\%) & 2,696 (4.78\%) \\
Researcher & 2,067 (4.59\%) & 256 (4.56\%) & 250 (4.37\%) & 2,573 (4.56\%) \\
Object & 2,009 (4.46\%) & 250 (4.45\%) & 256 (4.48\%) & 2,515 (4.46\%) \\
Proposer & 1,901 (4.22\%) & 217 (3.86\%) & 240 (4.20\%) & 2,358 (4.18\%) \\
Fault & 1,577 (3.50\%) & 191 (3.40\%) & 231 (4.04\%) & 1,999 (3.54\%) \\
Arg1 & 1,329 (2.95\%) & 164 (2.92\%) & 155 (2.71\%) & 1,648 (2.92\%) \\
Result & 1,302 (2.89\%) & 170 (3.03\%) & 166 (2.90\%) & 1,638 (2.90\%) \\
Arg2 & 1,250 (2.77\%) & 155 (2.76\%) & 152 (2.66\%) & 1,557 (2.76\%) \\
Concern & 1,066 (2.36\%) & 149 (2.65\%) & 144 (2.52\%) & 1,359 (2.41\%) \\
Extent & 998 (2.21\%) & 109 (1.94\%) & 133 (2.33\%) & 1,240 (2.20\%) \\
Dataset & 914 (2.03\%) & 115 (2.05\%) & 136 (2.38\%) & 1,165 (2.07\%) \\
Metrics & 794 (1.76\%) & 90 (1.60\%) & 92 (1.61\%) & 976 (1.73\%) \\
Finder & 695 (1.54\%) & 94 (1.67\%) & 74 (1.29\%) & 863 (1.53\%) \\
Reason & 26 (0.06\%) & 2 (0.04\%) & 0 (0.00\%) & 28 (0.05\%) \\
\midrule
\textbf{Total} & \textbf{45,076} & \textbf{5,618} & \textbf{5,717} & \textbf{56,411} \\
\bottomrule
\end{tabular}
\caption{Distribution of Argument Types across Train, Develop and Test Splits}
\label{tab:argument_distribution}
\end{table*}

\end{document}